\pdfoutput=1

\documentclass[11pt]{article}

\usepackage{acl}

\usepackage{times}
\usepackage{latexsym}

\usepackage[T1]{fontenc}

\usepackage[utf8]{inputenc}

\usepackage{microtype}

\usepackage{inconsolata}

\usepackage{easyReview}

\usepackage{tabularx}
\usepackage{booktabs}

\usepackage[most]{tcolorbox}

\usepackage[inline]{enumitem}

\usepackage{pifont}
\usepackage{amsmath}

\input{webis-paper-prefix}

\definecolor{cmark}{rgb}{0.2862, 0.5176, 0.4039}
\newcommand{\cmark}{\textcolor{cmark}{\ding{51}}}
\definecolor{xmark}{rgb}{0.7294, 0.2313, 0.2745}
\newcommand{\xmark}{\textcolor{xmark}{\ding{55}}}
\definecolor{neutral}{rgb}{0.9294, 0.6078, 0.2509}
\DeclareMathOperator*{\argmax}{arg\,max}
\newcommand{\neutral}{\textcolor{neutral}{\textbf{--}}}

\newcommand{\bspubtag}{
    \vspace*{-18.6cm}\hspace*{-0.5cm}
    {\fontsize{6}{8}\selectfont%
        \renewcommand{\arraystretch}{0.9}
        \begin{tabular}{l}
            Accepted to \\
            North American Chapter of the Association for Computational Linguistics (NAACL) 2025
        \end{tabular}
}}

\begin{document}
    
\title{Adaptive Prompting: Ad-hoc Prompt Composition for Social Bias Detection}

\author{
    Maximilian Spliethöver \textsuperscript{1},
    Tim Knebler \textsuperscript{1},
    Fabian Fumagalli \textsuperscript{2},
    Maximilian Muschalik \textsuperscript{3},
    \\
    \textbf{Barbara Hammer \textsuperscript{2},}
    \textbf{Eyke Hüllermeier \textsuperscript{3},}
    \textbf{Henning Wachsmuth \textsuperscript{1}}
    \\
    \textsuperscript{1}Leibniz University Hannover, Institute of Artificial Intelligence \\
    \textsuperscript{2}Bielefeld University, CITEC \\
    \textsuperscript{3}LMU Munich, MCML \\
    \tt \href{mailto:m.spliethoever@ai.uni-hannover.de}{m.spliethoever@ai.uni-hannover.de}
}

\maketitle

\begin{abstract}
Recent advances on instruction fine-tuning have led to the development of various prompting techniques for large language models, such as explicit reasoning steps.
However, the success of techniques depends on various parameters, such as the task, language model, and context provided. Finding an effective prompt is, therefore, often a trial-and-error process.
Most existing approaches to automatic prompting aim to optimize individual techniques instead of compositions of techniques and their dependence on the input.
To fill this gap, we propose an \emph{adaptive prompting} approach that predicts the optimal prompt composition ad-hoc for a given input.
We apply our approach to social bias detection, a highly context-dependent task that requires semantic understanding.
We evaluate it with three large language models on three datasets, comparing compositions to individual techniques and other baselines.
The results underline the importance of finding an effective prompt composition. Our approach robustly ensures high detection performance, and is best in several settings. Moreover, first experiments on other tasks support its generalizability.
\end{abstract}

    \bspubtag
    \vspace*{17.6cm}\hspace*{0.5cm}

    \section{Introduction}
\label{sec:introduction}

The development of instruction-tuned large language models (LLMs) has led to an increased interest in prompting techniques to augment inputs \cite{tian2024}. Whereas prompts for earlier generative language models consisted only of the input text and special tokens \cite{radford2018,raffel2020}, they may now encompass complex and explicit task instructions with ancillary information. Among others, successful prompting techniques include personas \cite{liu2024}, in-context demonstrations \cite{liu2023,dong2024}, and reasoning steps \cite{wei2022}.

\bsfigure{composition-example}{Exemplary excerpt of a prompt composition for social bias detection. While certain techniques might benefit detection performance (say, those with green squares), others might not (red squares). Some parts always need to be present (blue squares). A full prompt example is shown in Figure~\ref{fig:example-prompt}.}

The success and applicability of prompting techniques does, however, depend on several parameters, including the target task, the language model and its size, %
and the context provided \cite{brown2020,schick2021,mosbach2023,dong2024,arroyo2024}.
Finding an effective prompt is, therefore, often still a time-consuming process that needs re-evaluation should any of the parameters change. Recent automated prompting methods either address the lexical aspect by finding the best formulation \cite{honovich2023} or manipulate the latent space \cite{liu2022a}. Most automatic methods, however, focus on optimizing a single technique and do not consider a composition of techniques that could take advantage of their individual strengths.

The example in Figure~\ref{composition-example} visualizes a prompt composition of several techniques. While a \emph{definition} provides a theoretical background, \emph{in-context demonstrations} clarify its application. As recent works indicate that LLMs cannot attend equally to all available information \cite{liu2024a,plepi2024}, simply using more techniques may add noise and reduce performance. Finding an optimal prompt composition is, therefore, desirable.

To this end, we propose an \textit{adaptive prompting} approach to predict the optimal composition of discrete prompt techniques \emph{ad-hoc}, i.e., for each input instance. First, an encoder model learns to predict \textit{optimal compositions} based on a pool of individual prompting techniques. Consecutively, the approach predicts the best composition for each instance and prompts the LLM accordingly.

We evaluate adaptive prompting for five techniques and their compositions for the task of social bias detection. The task requires semantic understanding and world knowledge \cite{zhou2023a}, likely benefitting from using multiple prompting techniques, making it a good candidate to evaluate prompt compositions. To better understand the importance of each technique and their second-order interactions, we further conduct a Shapley interaction analysis \cite{fumagalli2023}.

We test our adaptive prompting approach on three social bias datasets for three open-weight LLMs, namely Mistral (7B) \cite{jiang2023}, Command-R (35B) \cite{cohereforai2024}, and Llama 3 (70B) \cite{dubey2024}. We compare to baselines within and across datasets, including a fine-tuned model and a composition ensemble. The results suggest that our approach robustly ensures high classification performance; in many cases, it even outperforms all baselines and fixed compositions. Moreover, follow-up experiments on three other NLP tasks stress the generalizability of our approach beyond social bias.

To summarize, our main contributions are:
\begin{itemize}
\setlength{\itemsep}{0pt}
    \item We present a novel discrete prompt optimization approach to predict the optimal prompt composition for a given model and input. It improves over several baselines and individual prompting techniques on selected datasets.
    \item We evaluate the utility of generative LLMs and prompting for social bias detection for three state-of-the-art LLMs on three datasets and with five prompting techniques.
    \item We provide insights into the performance and interaction of prompting techniques, finding that well-performing techniques can also interact negatively when used with others.%
\footnote{Code at: \url{https://github.com/webis-de/NAACL-25}.}
\end{itemize}

    \section{Related work}
\label{sec:related-work}

Discrete prompts for LLMs have been evaluated for various tasks and applications. For example,  \citet{zamfirescu-pereira2023} and \citet{arroyo2024} investigate how sub-optimal prompts affect outputs and \citet{hida2024} study how different prompts influence social bias exhibited by LLMs.

Several techniques have been proposed to optimize discrete prompts. Popular techniques include personas for perspective taking \cite{sheng2021,xu2023,liu2024}, in-context demonstrations to provide application examples \cite{dong2024}, and reasoning steps that divide the tasks into sub-tasks \cite{wei2022}. In this work, we do not consider single prompting techniques but rather evaluate prompt compositions to take advantage of several techniques. Some related works also explore the effect of combining techniques \cite{stahl:2024}, aiming to find the best composition on average. In contrast, we aim to find the optimal composition on each text.

Among existing approaches to automatic prompt optimization, continuous prompt optimization learns to adjust the latent space \cite{li2021b,liu2022a}, whereas several studies generate discrete optimized instructions from task examples, to easier adapt to unseen data \cite{zhou2023,honovich2023,ha2023} or new LLMs \cite{memon2024}. Other works focus on iteratively optimizing prompts \cite{zhang2022,shum2023,tian2024} or predicting the suitability of prompts  \cite{yang2024}.
Instead of optimizing the latent space or single techniques, we propose to automatically find optimal prompt compositions for unseen inputs.

Related to the idea of prompt compositions, \citet{khattab2023} propose a framework to optimize the use of multiple techniques by training a parameterized model that automatically optimizes the prompt. We do not optimize techniques but learn to predict the optimal composition of techniques for a given setting. We further analyze the importance of each technique using Shapley values.

While several other studies aim to detect social bias in text corpora \cite{spliethover2020,asr2021,toroisaza2023,derner2024}, we aim to identify bias in single text instances, similar to \citet{schick2021,spliethover2024,powers2024}, by optimizing prompt compositions.

\hwfigure{approach}{The three steps of our adaptive prompting approach: (1)~Bias labels are collected for all considered prompt compositions. (2)~A model is trained on the collected labels to predict the optimal composition for any given text. (3)~Given an unknown text, the model is applied to predict and use the optimal prompt composition for that text.}

Detecting social bias reliably requires understanding of social language and pragmatics to interpret the implications of text \cite{choi2023}. \citet{hovy2021} identify seven factors of language (e.g., receiver information) to successfully model social aspects. \citet{choi2023} and \citet{zhou2023a} design social language benchmarks and find that LLMs are still limited in this regard. In this work, prompt compositions enable the inclusion of social aspects (e.g., receiver information with persona prompts) and can, therefore, provide helpful context for social language tasks.

    \section{Approach}
\label{sec:method}

We evaluate instruction-tuned LLMs to identify social bias using compositions of prompting techniques. Specifically, we propose a three-step approach to predict the optimal composition ad-hoc per input text. The process is illustrated in Figure~\ref{approach}.

The first step is to create prompt compositions that represent order-sensitive combinations of one or more prompting techniques. We prompt an LLM with each of these compositions to collect social bias classification labels. In the second step, we use the collected labels to train an adaptive prompting model that finds the optimal prompt composition for a given text instance. Aside from the main task, we also conduct a Shapley value analysis to determine the importance of prompting techniques. We then apply adaptive prompting to unknown texts to find the optimal prompt composition for each.

\subsection{Composition and Label Collection}

We start by collecting bias label predictions for all possible prompt compositions, emanating from our base composition, a set of prompting techniques, and constraints for their ordering and compatibility.

\paragraph{Base Composition}
As the minimal prompt to solve the task, we consider the \emph{task description} and \emph{input instance} to be always present (cf. Figure~\ref{composition-example}).

\paragraph{Ordering and Compatibility Constraints}
Not all orderings and combinations of prompting techniques create meaningful prompts. For example, if both a \emph{definition} (e.g., of social bias) and \emph{in-context demonstrations} (e.g., a biased text) are present, demonstrations make sense after the definition only. To ensure meaningful prompts, we pre-define a general ordering. Further, variants of the same technique (e.g., demonstrations sampled randomly or by similarity) should not appear together in the same composition; they are mutually exclusive.

\paragraph{Prompting Techniques}
Let a set of $n \geq 1$ prompting techniques, $T = \{t_1,\dots,t_n\}$, be given with fixed order $t_i$ before $t_j$, if $i<j$. We distinguish techniques with one variant, $T_{1}$, and with multiple variants, $T_{2}$, that is, $T = T_{1} \sqcup T_{2}$, where each $t \in T_{2}$ has $\vert t \vert$ variants. Concretely, each technique in $T_1$ may or may not be used, and any or none of the techniques in $T_2$ may be used.
Let the set of distinct compositions be denoted as $C$. Then, the number of compositions is
\begin{equation}
    \label{eqn:number-of-compositions}
    \vert C \vert = 2^{ \vert T_1 \vert} \cdot \prod_{t \in T_2} (\vert t \vert + 1).
\end{equation}

\subsection{Composition Prediction and Selection}
\label{sec:composition-prediction}

As visualized in the second step of Figure~\ref{approach}, we use the results of the label collection described above to train a prompt composition \emph{prediction model}. The model is then used for \emph{adaptive prompting} (Step~3).

\paragraph{Prediction Model}
The model is an encoder model with a regression head that is fine-tuned on the collected social bias labels to predict the \textit{optimal} prompt composition
$c_o$ from the set of possible compositions $C$. Here, \textit{optimal} refers to the composition with the highest predicted likelihood to generate a correct bias label for an input.
We formulate the prediction of the optimal prompt composition as a regression problem using a sigmoid output layer, followed by binary cross-entropy loss \cite{ridnik2021,grivas2024}. The model's output is a $|C|$-dimensional vector, $\hat{y}$, in which each value represents an independent likelihood estimation for one of the $|C|$ compositions to be optimal.%
\footnote{We refrain from a multi-class setup, where the optimal composition is determined over a probability distribution spanning all compositions to avoid a few dominant compositions from possibly being preferred over others consistently.}

\paragraph{Adaptive Prompting}
Given an unknown text, the composition with the highest likelihood is assumed to be an optimal prompt composition $c_o$, where $o := \argmax(\hat{y})$. Thereby, we adaptively select a prompt depending on the text at hand.

\subsection{Shapley-Based Composition Analysis}

Analyzing the impact of individual prompting techniques and their interactions is crucial to evaluate outputs of the adaptive prompting model. To gain these insights, we rely on Shapley values  \cite{shapley1953}, modeling the predictive performance across all possible compositions as a cooperative game:

\paragraph{Prompt Composition Game}
Given the set of techniques $T$, let $\nu: 2^T \to \mathbb{R}$ be the performance of the techniques $T_c \subseteq T$ of a composition $c$:
\begin{equation}\label{def-prompt-composition}
    \nu(c) := \lambda(y,\hat y_c)
\end{equation}
Here, $\lambda$ is a performance metric, $y$ are the ground-truth bias labels, and $\hat y_c$ the predictions of $c$.
For techniques in $T_2$, a specific choice must be fixed.

We compute one Shapley value (SV) for each prompting technique $t \in T$, which provides contribution values $\phi(t)$. The SV quantifies the impact of a prompting technique across all possible compositions.
Beyond individual contributions, we compute pairwise Shapley interactions
\cite{Lundberg.2020} that additionally assign contributions to all pairs of techniques. Shapley interactions (SIs) reveal \emph{synergies} and \emph{redundancies} among prompting techniques, capturing the behavior of the game with greater fidelity \cite{Tsai.2022,fumagalli2024kernelshapiq}. Akin to Shapley-based feature or data selection \cite{Rozemberczki2022}, we select optimal compositions based on SVs and SIs.

    \section{Experiments}
\label{sec:experiments}

In this section, we detail the adaptive prompting experiments that we carried out for the task of social bias detection on three datasets with three state-of-the-art instruction-tuned LLMs. We selected five common prompting techniques that fit the task, which we combine to create prompt compositions. We evaluate our approach against own and related-work baselines, both within and across datasets.

\subsection{Task}
Social bias detection describes the task of identifying texts that induce bias against a particular social group, through offensive language, stereotypes \cite{nadeem2021}, power dynamics \cite{sap2020,zhou2023a}, or similar. More context can support the bias detection, making it a well-suited task to evaluate prompt compositions.

In particular, the task requires knowledge about the state of the world to understand implicit biases and dynamics \cite{hovy2021,zhou2023a}. Prompting techniques such as in-context demonstrations can show how to apply knowledge acquired during pre-training to identify implicit biases. Furthermore, what is considered a biased text can vary, e.g., based on the target application. Including definitions of bias for and deducting reasoning steps could help to clarify the context and make the predictions more reliable. Lastly, whether a text is considered biased partly depends on the receiver. Instructing the LLM to assume a specific persona can clarify the evaluating perspective.

\subsection{Data}
\label{sec:experiments:data-llms}
To cover multiple aspects of social bias, we evaluate prompt compositions and their predictability on three datasets covering diverse intentions and target applications. In the following, we briefly describe each dataset and how we derive binary bias labels (preprocessing and prompt details in Appendix~\ref{sec:appendix-experimental-details}).

\paragraph{StereoSet}
The StereoSet corpus \cite{nadeem2021} serves the evaluation of stereotypes in generative models. Here, stereotypes are ``over-generalized [beliefs] about a particular group of people'' \cite{nadeem2021}. The data consists of scenarios and target groups that can be combined to create \textit{stereotypical} (biased), \textit{anti-stereotypical}, and \textit{meaningless} (both not biased) texts.

\paragraph{SBIC}
The Social Bias Inference Corpus (SBIC) \citet{sap2020} is intended to model multiple aspects of social bias explicitly. We use the \emph{implicit bias} aspect as the target label, which indicates text that are offensive towards a specific social group.

\paragraph{CobraFrames}
The CobraFrames corpus \cite{zhou2023a} captures the social and situational context of biased statements. Among others, it captures bias as implicit power dynamics or stereotypes between the speaker and the listener. We follow \citet{zhou2023a} in converting the \textit{offensiveness} label into binary bias representations.

\subsection{Prompting Techniques}
\label{sec:experiments:prompts}
We select five common prompting techniques to investigate potential benefits of prompt compositions over single techniques for social bias detection. We focus on discrete prompting techniques as opposed to continuous methods, such as prefix-tuning \cite{li2021b} or p-tuning \cite{liu2022a}. Notice, though, that our adaptive prompting is applicable to arbitrary techniques in principle.

Since we evaluate prompt compositions across three datasets with different structures and definitions of social bias, specific content parts of the prompts are adjusted to better align with the scenario of each dataset. In the following, we briefly describe each technique, including dataset alignments and mutations. See Appendix~\ref{sec:appendix-experimental-details} for details on their lexical representations and a full example.

\paragraph{Personas}

This technique aims to instruct the model to follow consecutive instructions from the perspective of a specific persona \cite{thoppilan2022,deshpande2023}. Among other use cases, personas are used to build translation systems \cite{he2024} and dialogue agents \cite{thoppilan2022,xu2023}, or to investigate biases in pre-trained LLMs \cite{beck2024}.

For social bias detection, a persona can help clarify the perspective from which a given text is being judged \cite{giorgi2024}. Since the intentions and goals vary across datasets, we expect that a persona prompt can clarify the setting of the task. In our experiments, we aim to formulate the persona description as close as possible to the scenario envisioned for each dataset. Similar to \citet{xu2023}, we seek to minimize positionality bias.

\paragraph{Definitions}

Including a definition of social bias can be seen as an extension of the task description to further specify its subject. This is supposed to make the interpretation of social bias in the respective dataset explicit, which is otherwise learned implicitly only in a supervised learning setup. Recent research has found that such definitions can increase the prediction performance in low-resource settings \cite{elsner2023}.

If the authors provide an explicit definition of bias, we reuse it. Otherwise, we manually derive a definition from available information.

\paragraph{In-context Demonstrations}
Known also as few-shot examples, this technique is a form of in-context learning that ``allows language models to learn tasks given only a few examples'' \cite{dong2024}. In-context demonstrations are provided during inference as part of the prompt, unlike traditional fine-tuning where model parameters are optimized in a supervised learning phase \cite{mosbach2023}. In-context demonstrations have been shown to improve results on various target tasks \cite{zamfirescu-pereira2023,dong2024}.

While seemingly simple, this technique entails several points of variation, including the \emph{number} and \emph{selection} of examples \cite{liu2022,zhang2022,levy2023,bertsch2024,dong2024} and their \emph{ordering} \cite{lu2022,shum2023,liu2024a}. To keep the experiments conceivable, we use one demonstration per bias type, adjust the number of similar demonstrations accordingly, and do not evaluate the ordering, but we focus on three common variations to select demonstrations:
\begin{itemize}
\setlength{\itemsep}{0pt}
    \item \emph{Random.} We pseudo-randomly select training instances, which are used as demonstrations for all instances in the test split.
    \item \emph{Similarity.} For each instance in the test split, we select the most similar instances from the training split as demonstrations, using SBERT \cite{reimers2019}.
    \item \emph{Category.} We select instances that cover all bias types covered in a dataset, as diversifying demonstrations has been shown to aid prediction \cite{levy2023,zhang2022}.
\end{itemize}

\paragraph{Directional Stimulus}
Directional stimuli \cite{li2023} describe the technique to include instance-specific hints and are meant to guide the LLM. We include a list of dataset-specific bias types that could be present in the text instance.

\paragraph{Reasoning Step Instructions}
Initially intended for ``arithmetic, commonsense, and symbolic reasoning tasks'' \cite{wei2022}, the main idea of this technique is to decompose the given task into smaller, more approachable sub-tasks \cite{dong2024}. Reasoning step instructions have been applied to various tasks and can lead to improvements in prediction performance \cite{wei2022}.

As the detection of social bias in texts can often naturally be decomposed into step-wise questions \cite{sap2020,zhou2023a}, we include reasoning steps as an additional technique. We evaluate both zero-shot and few-shot settings. To do so, we follow \citet{press2023} and \citet{zhou2022a} in formulating the reasoning steps as task- and data-specific sub-questions covering the aspects of social bias in the respective dataset before prediction. To ensure that all predefined reasoning steps are followed as intended, we separate the reasoning steps into multiple consecutive inference steps \cite{dong2024}, implemented as a practical chain-of-prompts pipeline \cite{zhou2022a}.

\subsection{Models}

We realize our \emph{adaptive prompting} approach for three different \emph{instruction-tuned LLMs} as follows.

\paragraph{Adaptive Prompting}
Given the five prompting techniques, we fine-tune a DeBERTA-v3-large encoder model \cite{he2023} to predict the optimal composition ad-hoc, i.e., for a given input, as detailed in Section~\ref{sec:method}. Since the prompting techniques include three variants of in-context demonstrations that not compatible, the model predicts probabilities for $2^{4} * (3 + 1) = 64$ compositions (cf. Equation~\ref{eqn:number-of-compositions}). The composition with the highest probability is then used for the social bias classification.

We train a adaptive prompting model on the train split of each dataset, optimize it on the validation split, and evaluate its performance on the test split. We further train one adaptive prompting model per combination of dataset and LLM. Each model is trained with five different pseudo-random seeds.

\paragraph{Instruction-tuned LLMs}
We generate predictions with three instruction-tuned open-weight LLMs. To reliably generate classification labels, we use constrained decoding \cite{beck2024}, limiting the output to binary labels. Our LLM selection aims to diversify architecture, pre-training data, and size, as the effectiveness of prompting techniques may depend such factors (details on each model can be found in Appendix~\ref{sec:appendix-experimental-details}):
\begin{itemize}
\setlength{\itemsep}{0pt}
\item
\emph{Mistral.} The smallest LLM is Mistral-7B-Instruct-v0.2 \cite{jiang2023} with seven billion parameters. Its architecture focuses on generation performance and inference speed.
\item
\emph{Command-R.} As medium-sized LLM, we use C4AI Command-R v01 \cite{cohereforai2024} with 35 billion parameters. At the time of writing, architectural details were not available.
\item
\emph{Llama 3.} The largest evaluated LLM is Meta Llama 3 \cite{dubey2024}, with 70 billion parameters. It builds on a dense Transformer architecture to allow for easier scaling.
\end{itemize}

\subsection{Baselines}
\label{sec:experiments:baselines}

In addition to \emph{random} and \emph{majority} predictors that serve as lower bounds, we evaluate the following baselines to gain a comprehensive overview of the composition capabilities of our approach:

\paragraph{Self-Diagnosis}
Self-Diagnosis \cite{schick2021} adopts a Q\&A setting and a GPT-2 XL model \cite{radford2019} to identify social bias.

\paragraph{DeBERTa fine-tuned}
We fine-tune a DeBERTa-v3-large model \cite{he2023} in a supervised learning setting. It provides a reference to compare inference-only approaches and prompt compositions to a more traditional learning setup.

\paragraph{Ensemble}
The ensemble returns the label that was predicted most often across compositions. It helps to assess the value of adaptive prompting, compared to simply relying multiple compositions.

\paragraph{Best on Val/Test}
The Best on Val baseline represents the best-performing composition on the validation split. It gives insights as to whether adaptive prompting can perform better than any single composition on the evaluated datasets. Best on Test does the same for the best test split composition; notice that this knowledge is \emph{not} given in practice.

\paragraph{Oracle}
This upper bound produces a correct prediction if \emph{any} of the prompt compositions predicts the correct label. The oracle represents the hypothetical best performance that can be achieved.

    \section{Results and Discussion}
\label{sec:results}

\bsfigure{performance-stereoset}{Social bias detection results on StereoSet~(others in Appendix~\ref{sec:appendix-extended-results}: Figures~\ref{performance-sbic}--\ref{performance-cobra}): Macro F$_1$-score of all prompt compositions with each LLM (baselines shown as vertical lines). Our adaptive prompting approach (\textit{Ours StereoSet}) outperforms all fixed compositions. \emph{Ours SBIC} and \emph{Ours Cobra} are trained on other datasets. The variance over all compositions (shown as box plots) indicates the LLMs' sensitivity to the prompt.
}

Figure~\ref{performance-stereoset} shows the results of the prompt composition evaluation on StereoSet (see Appendix~\ref{sec:appendix-extended-results} for results on SBIC and CobraFrames). In the following, we highlight and discuss findings indicating the benefit of prompt compositions. Furthermore, we find that \emph{instance-specific} compositions can perform better than any \emph{single} composition alone.

\hwfigure{network-stereoset}{Network plots of the shapley interactions for the three evaluated LLMs on StereoSet (others in Appendix~\ref{sec:appendix-sv-composition-analysis}: Figures \ref{network-sbic}-\ref{network-cobra}), revealing unique interaction structures among the models. Node size represents strengths of first-order interactions. Line width and translucency denote strengths of second-order interactions. Red color denotes positive interaction (increasing the performance), and blue color denotes negative interaction (decreasing the performance).}

\subsection{Impact of Prompt Composition}

The results highlight the potential benefit of using prompt compositions compared to individual techniques or the base composition (exemplified for StereoSet in Table~\ref{tab:techniques-results-comparison-stereoset}). In experiments on Stereo\-Set, prompt compositions outperform individual techniques across all LLMs. The same is true for SBIC, but not for CobraFrames. While single techniques can still perform better when negative interactions between techniques inside a composition exist, our experiments highlight the benefit of using prompt compositions when choosing its techniques correctly. This is supported by the Shapley interactions, showing several positive interaction between techniques. Since compositions perform consistently better in our experiments, the results suggest that the benefit of compositions over single techniques holds across LLM architectures and sizes.

Some techniques are, however, notably more often present in the best-performing prompt composition than others, highlighting their positive impact. Table~\ref{tab:composition-frequencies-stereoset}, Table~\ref{tab:composition-frequencies-sbic}, and Table~\ref{tab:composition-frequencies-cobra} show how often each composition was chosen by our approach. On StereoSet, for example, in-context demonstrations are included in compositions that perform best on the test set and the validation set across models. A similar pattern exists for SBIC and CobraFrames. This finding is further supported by Shapley values and interactions, which highlight the strong positive contributions of in-context demonstrations.

Instead of using prompt compositions that include a selected technique with positive impacts, we observe that no single composition performs best across all datasets and LLMs in our experiments. Adapting the composition to input and LLM automatically is thus a crucial endeavor.

\begin{table}
    \small
    \renewcommand{\arraystretch}{.95}
    \centering
    \setlength{\tabcolsep}{1.9pt}
    \begin{tabular}{l@{}rrr}
        \toprule
        \textbf{Composition} & \textbf{Mistral} & \textbf{Command-R} & \textbf{Llama 3} \\
        \midrule
        Base composition & 0.711 & 0.462 & 0.575 \\
        [.5em]
        Definition & 0.716 & 0.527 & 0.637 \\
        Directional stimulus & 0.662 & 0.584 & 0.566 \\
        Persona & 0.698 & 0.546 & 0.539 \\
        Reasoning steps & 0.697 & 0.509 & 0.610 \\
        Demonstrations: Random & 0.665 & 0.674 & 0.725 \\
        Demonstrations: Category  & 0.681 & 0.675 & 0.739 \\
        Demonstrations: Similar & 0.761 & 0.701 & 0.798 \\
        [.5em]
        Best on Test & 0.800 & 0.706 & 0.817 \\
        Best by Shapley values & 0.790 & 0.588 & 0.798 \\
        Best by Shapley interaction & 0.795 & 0.671 & 0.800 \\
        [.5em]
        Adaptive prompting & \textbf{0.809} & \dag{} \textbf{0.781} & \ddag{} \textbf{0.853} \\
        \bottomrule
    \end{tabular}

    \caption{Macro F$_1$-score of each individual technique and selected prompt compositions on StereoSet (others in Appendix~\ref{sec:appendix-extended-results}). Results marked in bold indicate the best score per LLM. \textit{Best on test} describes the compositions that perform best on the test set for each model, and the two rows the best compositions based on Shapley values and interactions. Adaptive prompting is significantly better than \textit{Best on test} (\dag{} for $p<.05$, \ddag{} for $p<.01$).}
    \label{tab:techniques-results-comparison-stereoset}
\end{table}

\subsection{Volatility of Composition Performance}

In line with previous research \cite{zamfirescu-pereira2023,errica2024,memon2024}, our results highlight the sensitivity of current LLMs towards changes in prompt composition and data. While all evaluated compositions perform better than the \emph{Self-Diagnosis} baseline, there is a performance gap between the best and the worst prompt composition for all three LLMs (see Figure~\ref{performance-stereoset}). This stresses the difficulty of choosing a prompt that performs consistent across models (e.g., for Llama~3, 0.817 macro F$_1$ with the best composition, 0.483 with the worst). Using the worst compositions even partly led to results below the \emph{random baseline} or \emph{majority baseline} for Command-R and Llama~3.

Furthermore, the distance of \emph{best test} to the median indicates that the best compositions elicit very different behavior in the model compared to the majority: For Command-R and Llama~3, there are disparities of 0.115 and 0.107 between the median (0.591 and 0.710) and \emph{best test} (0.706 and 0.817).

Due to this sensitivity of LLMs to prompts and input data, the composition that elicits the best Macro F$_1$ also varies across a dataset. For example, while the best composition to predict the social bias label on Stereo\-Set contains \emph{definitions}, \emph{in-context demonstrations}, and a \emph{persona} for Llama~3, it is not the optimal composition for all instances and LLMs. Relying on a single composition for a whole dataset can, therefore, affect performance in unforeseeable ways. This clearly underlines the benefit of choosing LLM- and input-specific prompt compositions with an adaptive prompting approach.

\subsection{Impact of Adaptive Prompting}

The performance of our approach is particularly visible on StereoSet and SBIC. Choosing prompt compositions adapted to the input instance produces more reliable social bias detection on StereoSet across all three LLMs, for example, boosting the F$_1$-score from 0.706 to 0.781 for Command-R. On SBIC, the performance varies. Still, our approach is at least competitive to the best composition with Mistral (0.792 for \emph{best test} vs.\ 0.790 for adaptive prompting) and outperforms it with Llama~3 (0.831 vs.\ 0.842). These results provide further support for the idea of selecting input-specific compositions.

Adaptive prompting also ensures the use of prompt compositions that outperform (or are competitive to) \emph{DeBERTa fine-tuned}, whereas, on all three LLMs, most compositions perform worse. For example, in Figure~\ref{performance-stereoset}, the median composition score is 0.698 with Mistral vs.\ 0.781 for DeBERTa. With Command-R (0.591), it is even closer to the \emph{random baseline} (0.497) than to DeBERTa (0.781).

\subsection{Shapley Prompt Composition Analysis}

The results of the Shapley-based analysis further support the benefit of adaptive prompt compositions and find strong interactions between several prompting techniques, exemplified in Figure~\ref{network-stereoset} for StereoSet (details in Appendix~\ref{sec:appendix-sv-composition-analysis}).

While making use of prompting techniques improves performance in general, simply adding all possible techniques to a composition does not consistently enhance performance compared to providing only a task description across settings. This highlights the benefit of using and prompt compositions adapting them to the input instance.

Furthermore, the Shapley-based results suggest that the selection of compositions requires empirical validation or optimization, as the best-on-test compositions never contain all techniques but rather a heterogeneous set. The heterogeneity of the compositions suggests the need for a more stringent mechanism in selecting the best compositions, such as learning a meta-composition prediction model (such as \emph{adaptive prompting}) or conducting a game-theoretic assessment.

Lastly, choosing the composition based on Shapley values instead of our Adaptive prompting improves performance compared to baselines where no additional information is used, i.e., using no technique or all techniques. Modeling the selection problem with Shapley interactions instead further improves the performance of composition choices over Shapley values for StereoSet.

\subsection{Encoder Evaluation}

To investigate the performance of the encoder model, we evaluate its ability to predict optimal compositions that result in correct classifications.

Furthermore, we evaluate the composition selection frequencies and how often each composition resulted in correct predictions. This method shows whether the encoder model simply overfits the training set and simply predicts the most common composition (further details in  Appendix~\ref{sec:appendix-extended-results}).

The results suggest that the encoder model indeed learns to select optimal compositions. While the encoder performs better in predicting optimal compositions for Llama~3 and worse for Command-R on StereoSet and SBIC, the results are more mixed on CobraFrames.

Furthermore, comparing composition prediction frequencies across datasets indicates that the encoder model does not overfit the training set. Given that no single composition results in notably more correct predictions in the training split of each corpus, there is also little incentive for the encoder model to overfit to a single composition.

For CobraFrames, however, the results suggest that the encoder model can likely not learn meaningful connections between the inputs and compositions. This behavior, in turn, likely causes the comparably low performance of our adaptive prompting on CobraFrame. Adaptive prompting does, however, still avoid the risk of choosing a very ineffective prompt on CobraFrames.

Overall, the optimal compositions chosen by our encoder model show promising results. However, larger encoder models might be able to encode dependencies between text instances and prompt composition performance even better and deal with complex inputs more reliably.

\subsection{Adaptive Prompting across Datasets}

\emph{Ours Cobra} and \emph{Ours SBIC} in Figure~\ref{performance-stereoset} have been trained on the other datasets. The results  suggest that adaptive prompting does not perform as well across datasets compared to in-dataset training. This issue might be partially related to the sensitivity of LLMs to the prompt discussed above, that is, knowledge of prompt compositions may not be transferable across datasets. A potential reason for performance disparities could be the domain and format of the input text. While instances in StereoSet and CobraFrames are curated and contain sentences with a clear structure, those in SBIC come from online forums with noisy elements.

In some settings, though, our approach seems to generalize across datasets to some extent; for example, \emph{Our Cobra} performs above the median score on all three LLMs on StereoSet. This finding supports our hypothesis that the input data format can be relevant for predicting prompt compositions.

\subsection{Adaptive Prompting for Other Tasks}

To further validate the value of adaptive prompting, we trained and evaluated our approach on three additional tasks: sentiment analysis, natural language inference, and question answering (see Table~\ref{tab:other-tasks-results} in Appendix~\ref{sec:appendix-extended-results}). While the gains over single compositions are smaller than for social bias detection, adaptive prompting performs significantly better than the \emph{base composition} in all cases and generates better predictions than the \emph{Best on Val} baseline on all three tasks (recall that \emph{Best on Test} is more theoretical, as it cannot be found in practice).

We conclude that the idea of composing prompts ad-hoc dependent on the input instance (as realized for the first time in our approach) may have potential for many NLP tasks. Further investigations are left to future research.

    \section{Conclusion}
\label{sec:conclusion}

In this paper, we have introduced the notion of prompt compositions, that is, combining multiple prompting techniques to improve LLM performance. We have further proposed an adaptive prompting model that learns to predict optimal prompt compositions ad-hoc, based on the input instance in the context of social bias detection.

Through extensive experiments and a Shapley analysis, we have provided insights into the utility and importance of several prompting techniques for the given task. We find that the benefit of each technique and composition notably depends on the input and the LLM used, highlighting the need for automated systems to optimize prompt. We show that our adaptive prompting approach can improve upon single compositions on selected datasets.

In future work, we seek to work on technique- and task-agnostic approaches to find optimal prompt compositions and do so more efficiently. We hope that our work contributes towards fairer NLP through better social bias detection systems and enables research on using LLMs more efficiently through better prompting techniques.

    \section*{Limitations}
\label{sec:limitations}

For a focused study, we have exclusively modeled the detection of social bias as a binary classification setting. While we have considered multiple facets and settings of social bias by evaluating three datasets that employ diverse settings and definitions, the decision of whether a text elicits social bias or not is often more sophisticated than a binary answer. Our approach might not be applicable to settings requiring more nuanced decisions, but it still can support debiasing decisions or output explanations within respective NLP systems. It should, therefore, serve as a stepping stone to better and more inclusive systems.

As already explained in the main part of the paper, the proposed experiments are exhaustive, and their computational requirements depend on the number of  prompting techniques included, with a near-exponential growth in the inference steps required during training. Future work may aim to abstract from the specific techniques and learn to predict compositions by approximating their importance. We hope that the publication of our experimental data and results pave the way for more efficient adaptive prompting approaches and serve as a training ground to evaluate their feasibility of lowering the computations required.

As our experiments focus on prompt compositions, we do not evaluate lexical variations of the techniques and use a single phrasing per dataset for each technique. While lexical variations can influence the predictions of the model, we think that the general benefit of adaptive prompting still holds for prompting techniques with different phrasing, as the method is independent of the lexical properties of the prompt and rather learns from its predictions.

Lastly, our experimental setting focuses on the task of social bias detection, and the insights presented should, therefore, be considered in this context. However, we think the results are transferable to other tasks in the sense that the benefit of compositions and automatic prediction holds across tasks. Such transfer of the presented approach may require adjustments though, as also discussed in Section~\ref{sec:results}. The task we have focused on further limits our selection of techniques. Including more target tasks in the evaluation could allow for a more diverse selection of techniques, but it also requires a technique-agnostic approach to selecting optimal compositions. We plan to address this aspect in the future.

    \section*{Ethical Considerations}
\label{sec:ethical}

We aim to contribute towards a better detection of social biases in texts using current LLMs, considering different aspects, definitions, and scenarios of social bias by including diverse datasets. While this can be seen as a starting point towards a more reliable social bias detection, it is not a comprehensive evaluation of potential real-world scenarios. Therefore, the developed and published tools and data are research artifacts that are not ready for production. We, therefore, see the possibility that, when applied in real-world scenarios, the systems developed might elicit a false sense of trust in texts regarding their level of social bias, for example, due to misclassifications.

Another noteworthy aspect of this study is the environmental footprint. As discussed above, our experiments are extensive and require many GPU hours to be conducted. We, therefore, contributed to the growing carbon footprint of LLMs. However, we are confident that the data gathered can contribute towards using fewer computational resources, as predicting a prompt composition is computationally efficient (i.e., inference with a pre-trained model) and avoids constant re-prompting to find the best prompt. Furthermore, we hope that the publication of models and data helps to avoid the need to redo such experiments in the near future.

    \section*{Acknowledgments}

This work has been supported by the Deutsche Forschungsgemeinschaft (DFG, German Research
Foundation) under project number TRR 318/1 2021 – 438445824 and the Federal Ministry of Education and Research (BMBF), Germany under the AI service center KISSKI (grant no. 01IS22093C). We thank the anonymous reviewers for their valuable feedback and suggestions. The writing of the experimental code was supported by ChaptGPT and GitHub Copilot.

    \bibliography{arr24-prompting-based-bias-detection-lit}

    \appendix
    \section{Dataset Details}
\label{sec:appendix-data-details}

While all three datasets model some aspect of social bias, each dataset has a different goal, such as modeling multiple aspects of social bias, clarifying contextual settings, or evaluating stereotypes in generative language models. This section, therefore, extends on Section~\ref{sec:experiments:data-llms} to detail each corpus and the steps taken to prepare the datasets used in our experiments, including SBIC, Stereoset, and CobraFrames.

Table~\ref{tab:dataset-statistics} provides an overview of the number of positive and negative text instances per corpus and split.

\paragraph{In-context Demonstrations}
For each corpus, we select the most similar instances based on the cosine similarity of the sentence embeddings. We use the all-mpnet-base-v2 model from the Sentence-BERT library \cite{reimers2019} to generate the embeddings. For similarity-based demonstrations, we include the same number of demonstrations as for the category-based demonstrations to keep the experiments as comparable as possible. We thus include seven, four, and eleven demonstrations for the SBIC, StereoSet, and CobraFrames corpus, respectively.

\subsection{StereoSet Corpus}
\paragraph{Preprocessing}
For our experiments, we use the intersentence dataset of the StereoSet corpus. We construct instances with binary annotations by combining the provided context with each of the three assigned targets from the original dataset (labeled as \textit{anti-stereotype}, \textit{stereotype}, and \textit{unrelated}). Each context-sentence pair is then treated as a new instance.

We assign binary bias labels to the instances, where \textit{stereotype} sentences represents a bias text instance, while \textit{anti-stereotype} and \textit{unrelated} sentences represent a non-biased instance.

This method results in 6,369 instances. We pseudo-randomly split the data into training, validation, and test set with an 80/10/10 ratio.

\paragraph{Prompt Adjustments}
Since \citet{nadeem2021} focus on stereotypes prevalent in the USA (i.e., the selection of crowd workers was explicitly restricted to the USA), we include this information about the geographical target region in the definition of bias. Furthermore, we derive the reasoning steps from the data annotation guidelines and instruct the model to follow the persona of an annotator.

\subsection{Social Bias Inference Corpus}
\paragraph{Preprocessing}
To enhance data quality, we remove duplicate texts from the dataset and preprocess the remaining texts by removing characters, such as newlines, html entities, unicode characters, and multiple whitespaces. Since the original bias implication labels (\texttt{hasBiasedImplications}) in the SBIC dataset seem to be formatted incorrectly, with positive (label 1) indicating no bias and a negative (label 0) indicating bias. We, therefore, switch the labels so that a positive label indicates the presence of bias and a negative label indicates no bias.

We utilize the original validation and test splits provided by \citet{sap2020} with 4,666 and 4,691 samples, respectively. As explained in Section~\ref{sec:experiments} we sub-sample the training split, pseudo-randomly sampling 5,000 instances. The sampling is done in a stratified way that ensures a uniform distribution of the bias categories, as well as the bias label.

\paragraph{Prompt Adjustments}
Since the corpus combines data collected from micro-blogging platforms and forums, we instruct the model to assume the persona of a social media safety officer whose task is to flag biased social media posts. We further align the reasoning steps to the annotation questionnaire presented to the crowd workers, as published by \citet{sap2020}.

\subsection{CobraFrames Corpus}
\paragraph{Preprocessing}
Following the approach of \citet{zhou2023a}, we construct instances by concatenating the speaker identity, listener identity, speech context, and statement (available annotations for each instance) in a sentences as follows: ``This is a conversation between \texttt{[speakerIdentity]} and \texttt{[listenerIdentity]} in \texttt{[speechContext]}: \texttt{[statement]}.''

To generate binary social bias annotations, we convert the offensiveness dimension into a binary format based on the presence of specific phrases (e.g., ``offensive'', ``microaggression'' or ``xenophobic''), again following the approach of \citet{zhou2023a}.

For our experiments, we utilize both CobraCorpus and CobraCorpus-CF. From CobraCorpus, we pseudo-randomly sample 2,000 instances each for the training and validation sets in a stratified way, maintaining the original distribution of bias categories and bias labels. The CobraCorpus-CF is used as an additional test set.

To align the target group annotations between CobraCorpus-CF and CobraCorpus, we compute sentence embeddings for each target group in both corpora using the sentence-transformers library \cite{reimers2019} and the \texttt{all-mpnet-base-v2} model. Subsequently, each instance in CobraCorpus-CF was assigned the label of the target group from CobraCorpus with the highest cosine similarity. We manually validate the correctness of this process on several instances. Target groups that appeared in fewer than five instances in the resulting CobraCorpus-CF were excluded. The final test split comprised 1,939 samples.

\paragraph{Prompt Adjustments}
Since the dataset is designed around situational contexts with speaker and listener parties, we instruct the model to assume the persona of a third party overhearing the utterance and knowing about the identity and background of the speaker and listener. To create reasoning steps, we first annotate the intent, the potential target minority, and the implied statement before generating the final bias label prediction.

\subsection{Dataset Subsampling}
Due to the extensive nature of our experiments (i.e., we need to predict a label for each instance $|C|$ times, once for each composition, across all seeds), we sub-sample the training and validation splits of the SBIC and CobraFrames datasets. The exact number of instances per split and label are shown in Table~\ref{tab:dataset-statistics}.

    \section{Experimental Details}
\label{sec:appendix-experimental-details}

\subsection{Instruction-tuned LLMs}
Below, we shortly describe the information available for each LLM included in our evaluation.

\paragraph{Mistral}
The smallest model we evaluate is the instruction-tuned variant of
Mistral-7B-Instruct-v0.2 \cite{jiang2023}. It is based on the transformer architecture and introduces several architectural changes that improve its generation performance and inference speed. It is trained with seven billion parameters and has a context size of 32 thousand tokens.%
\footnote{\url{https://huggingface.co/mistralai/Mistral-7B-Instruct-v0.2}}
The authors do not publish information about the models' training data and procedure.

\paragraph{Command-R}
As medium-sized LLMs, we include C4AI Command-R v01 \cite{cohereforai2024}. The LLM is trained with 35 billion parameters and has a context length of 128 thousand tokens. To the best of our knowledge, no details on the architecture and training procedure are available at the time of writing.

\paragraph{Llama 3}
Lastly, as the biggest LLM, we include a instruction-tuned variant of the Llama~3 model family, namely Meta-Llama-3-70B-Instruct \cite{dubey2024}. The LLM was pre-trained with 70 billion parameters on more than 15 trillion tokens and has a context length of eight thousand tokens. The instruction-tuning was achieved with a mix of supervised fine-tuning and Reinforcement learning with Human Feedback \cite{dubey2024}. Similar to the Mistral and Command-R model, the authors do not publish information about the training data and procedure but specify the training data cut-off as December 2023 \cite{dubey2024}.

\paragraph{Generating Binary Classification Labels}
While standard classification models predict a single label and distribute probabilities only over the specified labels, instruction-tuned LLMs generate fluent text that is only restricted by the text in their training data. Reliably generating classification tokens is thus a challenge as, in some cases, LLMs might also generate tokens other than the expected labels.

To alleviate this problem, we restrict the logits of the models to the labels using constrained decoding \cite{beck2024}, as implemented in the outlines%
\footnote{\url{https://github.com/outlines-dev/outlines}}
and vLLM%
\footnote{\url{https://github.com/vllm-project/vllm}}
libraries. We restrict the decoding to the tokens ``Yes'' and ``No''.%
\footnote{In a pilot study, we also experiment with other versions of the generated token, such as yes/no, y/n, and 1/0, but find that Yes/No to produce the best results.}
To still allow for the generation of reasoning steps that require generating more than the binary labels, we first generate the reasoning steps without any restrictions and only restrict the decoding for the final label prediction.

\paragraph{Full Example Prompt}
Figure~\ref{fig:example-prompt} shows an example of a full and unmodified prompt composition that includes all possible prompting techniques with similarity-based in-context demonstrations, as used in our experiments for StereoSet.

\subsection{Technical Inference Setup}
The instruction-tuned LLMs for the social bias detection are run on four A100-SXM4-80GB and 16 H100-SXM-80GB GPUs. To ensure efficient inference given the numerous prompt compositions, we use the vLLM library with dynamic batching alongside the outlines library for constrained decoding. Additionally, inference is parallelized across the different prompt compositions to accelerate the overall process. With this setup, the inference for Mistral across all three datasets was completed in 26 hours, while the inference required 73 hours and 140 hours, for Command-R and Llama~3, respectively.

\subsection{Significance Testing}
We test for significant improvements of the proposed Adaptive Prompting approach over the best individual composition (if Adaptive Prompting shows the best overall results), as indicated in Table~\ref{tab:techniques-results-comparison-stereoset}, Table~\ref{tab:other-tasks-results}, Table~\ref{tab:techniques-results-comparison-sbic}, and Table~\ref{tab:techniques-results-comparison-cobra}.

Since we have access to all per-instance predictions for all models, we employ a one-sided independent $t$-test to compute significance levels of potential improvements of the Adaptive Prompting approach over the best individual compositions. We compute the significance levels over the results of all five random seeds per model and approach combination. The distributions of individual results matched the $t$-test assumptions.

We test for the two common $p$-values $p<0.05$ and $p<0.01$.

    \section{Shapley-based Composition Analysis}
\label{sec:appendix-sv-composition-analysis}

To understand the relationships between the different prompting techniques of a composition, we conduct a game theoretic analysis based on the Shapley value (SV) and Shapley Interactions (SI).
We further use the results from this Shapley-based analysis to predict optimal compositions for each model and dataset.

\paragraph{Setup}
To gain further insights into the interplay of prompting techniques, we analyze the prompt composition games (cf. Equation~\ref{def-prompt-composition}) across three datasets and models, exploring all possible variants of in-context demonstrations (category, similarity, and random).
Specifically, the players in each game include the personas (per.), definitions (def.), the specified in-context demonstration variant (cat./sim./rand. dem.), reasoning step instructions (rea.), and directional stimulus (dir. stim.).

We evaluate the games on all $\vert 2^T \vert = 2^5 = 32$ compositions, measuring the macro F1 scores of the models on both the \textit{validation} and \textit{test} sets for each composition $S \subseteq T$.
Next, we compute exact SVs and pairwise SIs \cite{Bord.2023,Lundberg.2020} on the \emph{validation} set using the \texttt{shapiq}\footnote{\url{https://github.com/mmschlk/shapiq}} package \cite{muschalik2025}.

Similar to Section~\ref{sec:composition-prediction}, we use the SVs and SIs to predict an optimal composition for each setting based on the validation data.
We reconstruct all game values for each composition $S \subseteq T$ using the SVs and SIs to select the set of prompting techniques with the highest reconstructed macro F1 score.
Formally, we iterate over all $S \subseteq T$ to combine the individual SV or pairwise SI scores into an additive prediction of the game with
\begin{align*}
    \hat{\nu}^\text{SV}(S) := \sum_{i \in S} \phi_i^{\text{SV}}
    \text{\quad and\quad}
    \hat{\nu}^\text{SI}(S) := \sum_{\substack{L \subseteq S\\|L| \leq 2}} \phi_L^{\text{SI}}
\end{align*}
where $\phi^{\text{SV}}$ and $\phi^{\text{SI}}$ are the SV and SI scores, respectively.
We then compare the performance of this selected composition on the \emph{test} dataset against naive compositions (using all techniques or none) and the overall best-performing compositions.

\begin{figure*}
    \centering

    \includegraphics[width=1\linewidth]{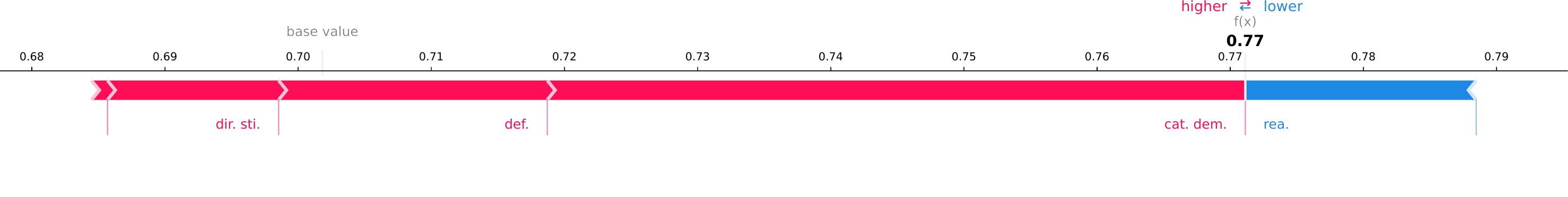}
    \\[-1em]
    \includegraphics[width=1\linewidth]{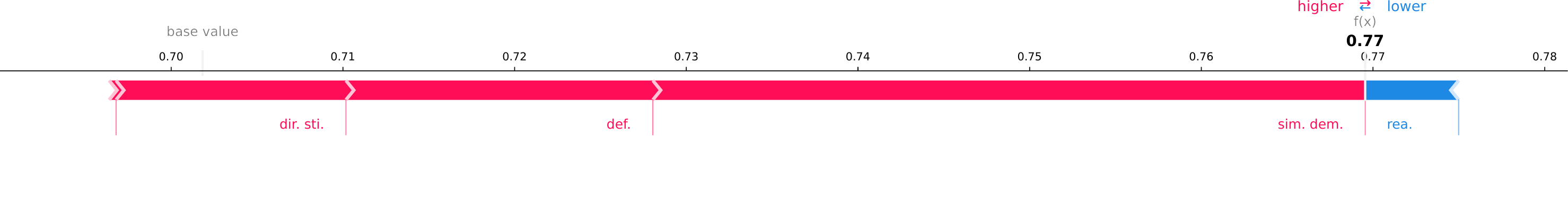}
    \\[-1em]
    \includegraphics[width=1\linewidth]{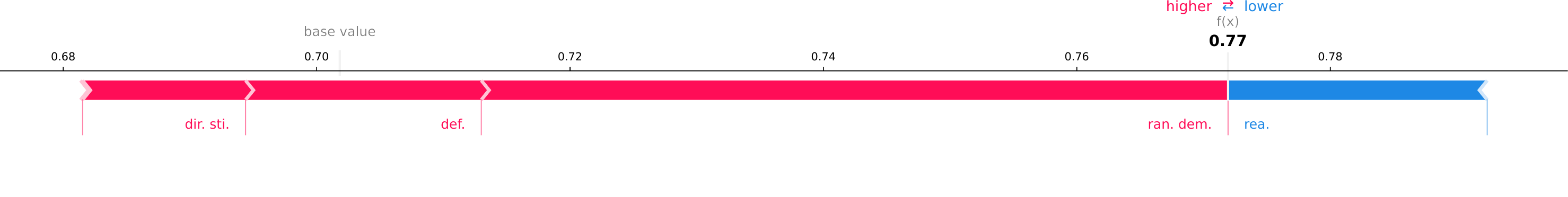}

    \caption{Force plots of Shapley values for three variants (top: category in-context demonstrations, middle: similar in-context demonstrations, bottom: random in-context demonstrations) of the composition game for the \emph{Mistral} model on \emph{Stereoset}. In all three settings the \textbf{in-context demonstrations are most influential}. Red color denotes positive attribution (increasing the performance), and blue color denotes negative attribution (decreasing the performance).}
    \label{fig_force_plots_stereoset}
\end{figure*}

\paragraph{Visualizing the SVs and SIs}
To visually investigate the SIs, we employ \emph{force} \cite{Lundberg.2017} and \emph{network} plots \cite{muschalik2024}.
Force plots, as presented in Figure~\ref{fig_force_plots_stereoset}, are commonly used to represent the SVs on a number line representing the prediction space.
On average, prompting techniques with a positive SV increase the performance of the models, and techniques with a negative value decrease the performance.
In the force plots this is represented by the positive techniques ``forcing'' the performance ``away'' from the performance of the empty composition $\nu(\emptyset)$ towards the performance of the full composition $\nu(T)$.
Additionally, the SIs indicate synergies (positive value) and redundancies (negative value) between prompting techniques \citep{fumagalli2024a}.
To illustrate second-order SIs among the individual prompting techniques, network plots, as depicted in Figure~\ref{network-stereoset}, Figure~\ref{network-sbic}, and Figure~\ref{network-cobra}, arrange the techniques in a circular layout and represent first-order and second-order interactions as nodes and edges, respectively.
The size of the nodes and edges represents the strength of the interactions, and the color denotes the direction (red increases performance, blue decreases performance).

\paragraph{Findings}
The results of the Shapley-based composition analysis are summarized in Table~\ref{tab:table-sv-composition-analysis} and Table~\ref{tab:table-si-selection}, as well as in Figure~\ref{fig_force_plots_stereoset}, Figure~\ref{network-stereoset}, Figure~\ref{network-sbic}, and Figure~\ref{network-cobra}.

Our results highlight a strong interaction between the different prompting techniques.
We present \emph{five} main findings.
\textbf{(1)} First, adding all possible prompting techniques to a composition does not consistently enhance performance compared to providing only a task description.
This is demonstrated in Table~\ref{tab:table-sv-composition-analysis}, where $\nu(T)$ (value of all compositions $T$) is not consistently higher than $\nu(\emptyset)$ (value of task description only) across all settings.
\textbf{(2)} Second, however, adding prompting techniques consistently improves performance, as all best-on-test compositions in Table~\ref{tab:table-sv-composition-analysis} consist of a non-empty set of techniques.
\textbf{(3)} Third, the selection of compositions requires empirical validation or optimization, as the best-on-test compositions \emph{never} contain all techniques but rather a \emph{heterogeneous} set.
The heterogeneity of the compositions suggests the need for a more stringent mechanism in selecting the best compositions, such as learning a \emph{meta-composition prediction model} or conducting a \emph{game-theoretic assessment}.
\textbf{(4)} Fourth, choosing the composition based on SVs improves performance compared to baseline conditions where no additional information is used, as SV compositions often outperform settings with either no prompting technique or all techniques.
\textbf{(5)} Fifth, modeling the selection problem with SIs, and thus with higher fidelity, substantially improves the performance of composition choices over SV-based selection for the StereoSet corpus, as summarized in Table~\ref{tab:table-si-selection}.

\hwfigure{network-sbic}{Network plots of the shapley interactions for the three evaluated LLMs on SBIC.}
\hwfigure{network-cobra}{Network plots of the shapley interactions for the three evaluated LLMs on CobraFrames.}

\begin{table*}
    \small
    \centering
    \setlength{\tabcolsep}{2.5pt}

    \begin{tabular}{@{}lllrrlrlrc@{}}
        \toprule
         &  &  &  &  & \multicolumn{2}{c}{\textbf{Best-on-Test Composition}} & \multicolumn{3}{c}{\textbf{SV Composition}} \\

        \cmidrule(l{2pt}r{2pt}){6-7}\cmidrule(l{2pt}){8-10}

        \textbf{Corpus} &  \textbf{Model} & \textbf{Variant} & \multicolumn{1}{l}{\textbf{$\nu(\emptyset)$}} & \multicolumn{1}{l}{\textbf{$\nu(T)$}}  & \textbf{Composition} & \multicolumn{1}{l}{\textbf{Score}} & \textbf{Composition} & \multicolumn{1}{l}{\textbf{Score}} & \multicolumn{1}{l}{\textbf{Best}} \\ \midrule

        \textbf{StereoSet\,\,} & \textbf{Mistral} & category & 0.711 & 0.682 & rea., def. & \textbf{0.722} & in-cont., rea. & 0.705 & \xmark \\
        &  & similar & 0.711 & 0.795 & in-cont., rea., def. & \textbf{0.800} & in-cont., rea., per. & 0.790 & \xmark \\
        &  & random & 0.711 & 0.705 & rea., def. & \textbf{0.722} & in-cont., rea. & 0.705 & \xmark \\

        & \textbf{Command-R} & category & 0.462 & 0.582 & in-cont., per. & \textbf{0.685} & in-cont., def., dir. sti., per. & 0.579 & \xmark \\
        &  & similar & 0.462 & 0.650 & in-cont., per. & \textbf{0.706} & in-cont., def., dir. sti., per. & 0.582 & \xmark \\
        &  & random & 0.462 & 0.652 & in-cont., per. & \textbf{0.677} & in-cont., def., dir. sti., per. & 0.588 & \xmark \\

        & \textbf{Llama~3} & category & 0.575 & 0.583 & in-cont., def. & \textbf{0.760} & in-cont., def. & \textbf{0.760} & \cmark \\
        &  & similar & 0.575 & 0.608 & in-cont., def., per. & \textbf{0.817} & in-cont., def., dir. sti. & \textbf{0.798} & \xmark \\
        &  & random & 0.575 & 0.495 & in-cont., rea. & \textbf{0.768} & in-cont., def. & \textbf{0.736} & \xmark

        \\\midrule

        \textbf{SBIC} & \textbf{Mistral} & category & 0.702 & 0.771 & in-cont., def., dir. sti., per. & \textbf{0.783} & in-cont., def., dir. sti., per. & \textbf{0.783} & \cmark \\
        &  & similar & 0.702 & 0.770 & in-cont., rea., def. & \textbf{0.772} & in-cont., def., dir. sti. & 0.758 & \xmark \\
        &  & random & 0.702 & 0.772 & in-cont., def., dir. sti. & \textbf{0.792} & in-cont., def., dir. sti. & \textbf{0.792} & \cmark \\

        & \textbf{Command-R} & category & 0.470 & 0.751 & in-cont., def. & \textbf{0.772} & in-cont., def., per. & \textbf{0.767} & \xmark \\
        &  & similar & 0.470 & 0.712 & in-cont., def., dir. sti. & \textbf{0.770} & in-cont., def. & \textbf{0.770} & \xmark \\
        &  & random & 0.470 & 0.724 & in-cont., def., per. & \textbf{0.788} & in-cont., def., per. & \textbf{0.788} & \cmark \\

        & \textbf{Llama~3} & category & 0.651 & 0.556 & in-cont., def., per. & \textbf{0.821} & in-cont., def. & \textbf{0.810} & \xmark \\
        &  & similar & 0.651 & 0.469 & in-cont., def., per. & \textbf{0.825} & def. & \textbf{0.788} & \xmark \\
        &  & random & 0.651 & 0.502 & in-cont., def., per. & \textbf{0.831} & in-cont., def. & \textbf{0.826} & \xmark

        \\\midrule

        \textbf{Cobra} & \textbf{Mistral} & category & 0.449 & 0.499 & rea., def. & \textbf{0.548} & in-cont., rea., def. & \textbf{0.522} & \xmark \\
        &  & similar & 0.449 & 0.532 & in-cont. & \textbf{0.604} & in-cont., def. & \textbf{0.604} & \xmark \\
        &  & random & 0.449 & 0.515 & rea., def. & \textbf{0.548} & in-cont., rea., def. & \textbf{0.544} & \xmark \\

        & \textbf{Command-R} & category & 0.535 & 0.633 & in-cont., rea., dir. sti., per. & \textbf{0.651} & in-cont., rea., def. & \textbf{0.633} & \xmark \\
        &  & similar & 0.535 & 0.641 & in-cont., rea., dir. sti. & \textbf{0.668} & in-cont., rea., def. & \textbf{0.639} & \xmark \\
        &  & random & 0.535 & 0.645 & in-cont., rea., def. & \textbf{0.654} & in-cont., rea., def., per. & \textbf{0.650} & \xmark \\

        & \textbf{Llama~3} & category & 0.461 & 0.536 & in-cont. & \textbf{0.599} & in-cont., def., dir. sti. & \textbf{0.599} & \xmark \\
        &  & similar & 0.461 & 0.376 & in-cont. & \textbf{0.605} & in-cont., def. & \textbf{0.594} & \xmark \\
        &  & random & 0.461 & 0.318 & in-cont., def. & \textbf{0.576} & in-cont., def. & \textbf{0.576} & \cmark \\

        \bottomrule
    \end{tabular}%

    \caption{Summary of Shapley Value-based composition selection on the test split for each corpus. For all three datasets, models and in-context demonstration \textit{variants} (category, similar, and random), the table depicts the F$_1$ scores (\textit{Score}) of the composition using no additional techniques ($\nu(\emptyset)$) and all remaining techniques ($\nu(T)$), the \textit{Best-on-Test Composition}, and the composition as determined by the Shapley Values (\textit{SV Composition}). Compositions improving over $\nu(\emptyset)$ and $\nu(T)$ are marked in bold.}
    \label{tab:table-sv-composition-analysis}
\end{table*}

\begin{table*}
    \footnotesize
    \renewcommand{\arraystretch}{.9}
    \centering
    \setlength{\tabcolsep}{1.4pt}

        \begin{tabular}{llllrclrcc}
            \toprule
            &  &  & \multicolumn{3}{c}{\textbf{SV Composition}} & \multicolumn{4}{c}{\textbf{SI Composition (2-SII)}} \\

            \cmidrule(r{2pt}){4-6}\cmidrule(l{2pt}){7-10}

            \textbf{Corpus} & \textbf{Model} & \textbf{Variant} & \textbf{Composition} & \textbf{Score} & \multicolumn{1}{c}{\textbf{Best}} & \textbf{Composition} & \textbf{Score} & \textbf{Best} & \textbf{SI > SV}

            \\\midrule

            \textbf{StereoSet} & \textbf{Mistral} & category & rea. & 0.705 & \xmark & -- \textcolor{gray}{task description only ($\emptyset$)} & \textbf{0.711} & \xmark & \cmark \\
            &  & similar & in-cont., rea., per. & 0.790 & \xmark & in-cont., rea., def.,   dir. sti., per. & \textbf{0.795} & \xmark & \cmark \\
            &  & random & in-cont., rea. & 0.705 & \xmark & in-cont., rea. & 0.705 & \xmark & \neutral \\

            & \textbf{Command-R} & category & def., dir. sti., per. & 0.579 & \xmark & def., per. & \textbf{0.669} & \xmark & \cmark \\
            &  & similar & in-cont., def., dir. sti., per. & 0.582 & \xmark & in-cont., def., per. & \textbf{0.671} & \xmark & \cmark \\
            &  & random & in-cont., def., dir. sti., per. & 0.588 & \xmark & in-cont., def., per. & \textbf{0.667} & \xmark & \cmark \\

            & \textbf{Llama~3} & category & def. & \textbf{0.760} & \cmark & def. & \textbf{0.760} & \cmark & \neutral \\
            &  & similar & in-cont., def., dir. sti. & \textbf{0.798} & \xmark & in-cont., def. & \textbf{0.800} & \xmark & \cmark \\
            &  & random & in-cont., def. & \textbf{0.736} & \xmark & in-cont., def. & \textbf{0.736} & \xmark & \neutral

            \\\midrule

            \textbf{SBIC} & \textbf{Mistral} & category & def., dir. sti., per. & \textbf{0.783} & \cmark & rea., def.,   dir. sti., per. & \textbf{0.771} & \xmark & \xmark \\
            &  & similar & in-cont., def., dir. sti. & 0.758 & \xmark & in-cont., rea., def.,   dir. sti., per. & \textbf{0.770} & \xmark & \cmark \\
            &  & random & in-cont., def., dir. sti. & \textbf{0.792} & \cmark & in-cont., def., dir.   sti. & \textbf{0.792} & \cmark & \neutral \\

            & \textbf{Command-R} & category & def., per. & \textbf{0.767} & \xmark & def., per. & \textbf{0.767} & \xmark & \neutral \\
            &  & similar & in-cont., def. & \textbf{0.770} & \xmark & in-cont., def., per. & \textbf{0.766} & \xmark & \xmark \\
            &  & random & in-cont., def., per. & \textbf{0.788} & \cmark & in-cont., def., per. & \textbf{0.788} & \cmark & \neutral \\

            & \textbf{Llama~3} & category & def. & \textbf{0.810} & \xmark & def. & \textbf{0.810} & \xmark & \neutral \\
            &  & similar & def. & \textbf{0.788} & \xmark & in-cont., def. & \textbf{0.821} & \xmark & \cmark \\
            &  & random & in-cont., def. & \textbf{0.826} & \xmark & in-cont., def. & \textbf{0.826} & \xmark & \neutral

            \\\midrule

            \textbf{Cobra} & \textbf{Mistral} & category & rea., def. & \textbf{0.522} & \xmark & rea., def. & \textbf{0.548} & \cmark & \cmark \\
            &  & similar & in-cont., def. & \textbf{0.604} & \xmark & rea., def. & \textbf{0.548} & \xmark & \xmark \\
            &  & random & in-cont., rea., def. & \textbf{0.544} & \xmark & in-cont., rea.,   def. & \textbf{0.544} & \xmark & \neutral \\

            & \textbf{Command-R} & category & rea., def. & \textbf{0.633} & \xmark & rea., per. & \textbf{0.648} & \xmark & \cmark \\
            &  & similar & in-cont., rea., def. & 0.639 & \xmark & in-cont., rea., dir.   sti., per. & \textbf{0.660} & \xmark & \cmark \\
            &  & random & in-cont., rea., def., per. & \textbf{0.650} & \xmark & in-cont., rea., dir.   sti., per. & \textbf{0.642} & \xmark & \xmark \\

            & \textbf{Llama~3} & category & def., dir. sti. & \textbf{0.599} & \xmark & def., dir.   sti., per. & \textbf{0.573} & \xmark & \xmark \\
            &  & similar & in-cont., def. & \textbf{0.594} & \xmark & in-cont., def., dir.   sti. & \textbf{0.574} & \xmark & \xmark \\
            &  & random & in-cont., def. & \textbf{0.576} & \cmark & in-cont., def. & \textbf{0.576} & \cmark & \neutral \\

            \bottomrule
        \end{tabular}%

    \caption{Summary of Shapley Interaction-based composition selection. For all three corpora, models and in-context demonstration \textit{variants} (category, similar, and random) games, the best composition and its F1 score (\textit{Score}) on the test split are shown for the composition selected with Shapley Values (\textit{SV Composition}) and Shapley Interactions (\textit{SI Composition}). Compositions improving over $\nu(\emptyset)$ and $\nu(T)$ (cf. Table~\ref{tab:table-sv-composition-analysis}) are marked in bold. Notably for StereoSet, compositions selected via Shapley interactions always improve or stay the same compared to compositions selected via the Shapley values.}
    \label{tab:table-si-selection}
\end{table*}

    \section{Extended results}
\label{sec:appendix-extended-results}

\subsection{Encoder model evaluation}

To investigate the raw performance of the adaptive prompting model in predicting prompt compositions ad-hoc based on the input text, as detailed in Section~\ref{sec:method}, we evaluate its ability to predict a composition that results in a correct classification (i.e., the optimal composition). This allows for a more direct view at the performance of the encoder model chosen for the approach.

Since our primary interest is an encoder model that is able to predict a composition that produces a correct classification for a given text instance and LLM, we consider all such compositions to be correct predictions of the encoder model ($\#correct\_predictions$). We then simply divide this number by the total number of instances ($\#instances$) in the dataset to calculate a ratio of correct predictions over the full dataset (i.e., $\frac{\#correct\_predictions}{\#instances}$). Like other classification metrics, the score range is $[0,1]$, where $1$ represents the best score. The results are shown in Table~\ref{tab:composition-prediction-performance-results}.

Furthermore, Table~\ref{tab:composition-frequencies-stereoset}, Table~\ref{tab:composition-frequencies-sbic}, and Table~\ref{tab:composition-frequencies-cobra} show the frequencies of how often the adaptive prompting approach chose a specific composition as the optimal composition and how often each composition produced a correct prediction for each model on the train dataset. All frequencies are averaged over five random seeds. This additional data is useful to evaluate, whether the encoder model overfits on the training dataset and simply predicts the most-common composition.

\bsfigure{performance-sbic}{Social bias detection results on SBIC: Macro F$_1$ of all prompt compositions for each LLM.}
\bsfigure{performance-cobra}{Social bias detection results on CobraFrames: Macro F$_1$ of all prompt compositions for each LLM.}

\subsection{Detailed Prompt Composition Results}
Figure~\ref{performance-sbic} and Figure~\ref{performance-cobra} show the boxplots for SBIC and CobraFrames, respectively.

Table~\ref{tab:techniques-results-comparison-sbic} and Table~\ref{tab:techniques-results-comparison-cobra} show a summary of the results, comparing individual techniques and adaptive prompting, similar to Table~\ref{tab:techniques-results-comparison-stereoset}.

Table~\ref{tab:compositions-performance-stereoset}, Table~\ref{tab:compositions-performance-sbic}, and Table~\ref{tab:compositions-performance-cobra} show the results for each evaluated composition on Stereoset, SBIC, and CobraFrames, respectively.

\subsection{Adaptive Prompting for Various Tasks}

Table~\ref{tab:other-tasks-results} shows the results of our adaptive prompting on three further tasks: For sentiment analysis, we use the Aspect Based Sentiment Analysis corpus \cite{pontiki2014}, also referred to as ABSA. For natural language inference, we use the e-SNLI corpus \cite{camburu2018}. Lastly, for question answer, we use the CommonsenseQA corpus \cite{talmor2019}.

We format the prompt as \texttt{<Q> question text <A> answer text}, for which the predicted label indicates whether the answer is correct, given the preceeding question. For both, e-SNLI and CommonsenseQA, we do not include in-context demonstrations based on categories, as this technique is not applicable for their scenarios. Otherwise, all results were retrieved using the same methodology and experimental setup presented in Section~\ref{sec:method} and Section~\ref{sec:experiments}. As LLM, we employ Mistral.

Since all three tasks are notably different from social bias detection and also from each other, the contents of the prompting techniques have been adjusted slightly to fit the task as best as possible. Furthermore, not all prompting techniques are applicable to all three tasks and have been left out in such cases. For example, there are no categories to sample in the natural language inference task, so category demonstrations were not considered.

\begin{table}
    \small
    \centering
    \setlength{\tabcolsep}{3pt}

    \begin{tabular}{lrrrrrr}
        \toprule
        & \multicolumn{2}{c}{\textbf{Training}} & \multicolumn{2}{c}{\textbf{Validation}} & \multicolumn{2}{c}{\textbf{Test}}  \\
        \cmidrule(l@{3pt}r@{3pt}){2-3}\cmidrule(l@{3pt}r@{3pt}){4-5}\cmidrule(l@{3pt}r@{3pt}){6-7}
        \textbf{Corpus} & \multicolumn{1}{l}{\textbf{Pos}}& \multicolumn{1}{l}{\textbf{Neg}} & \multicolumn{1}{l}{\textbf{Pos}}& \multicolumn{1}{l}{\textbf{Neg}} & \multicolumn{1}{l}{\textbf{Pos}} & \multicolumn{1}{l}{\textbf{Neg}} \\
        \midrule

        StereoSet & 1698 & 3397 & 213 & 424 & 212 & 425 \\
        SBIC & 2500 & 2500 & 1806 & 2860 & 1924 & 2767 \\
        CobraFrames & 1780 & 220 & 1779 & 221 & 1862 & 77 \\
        [.5em]
        ABSA & 1907 & 1009 & 240 & 123 & 249 & 111 \\
        ESNLI & 2500 & 2500 & 500 & 500 & 500 & 500 \\
        CommonsenseQA & 2500 & 2500 & 500 & 500 & 500 & 500 \\

        \bottomrule
    \end{tabular}

    \caption{The number of biased (\textit{Pos}) and not biased (\textit{Neg}) text instances per corpus and split.}
    \label{tab:dataset-statistics}
\end{table}

\begin{table}
    \small
    \centering
    \begin{tabular}{l@{}rrr}
        \toprule
        \textbf{LLM} & \textbf{StereoSet} & \textbf{SBIC} & \textbf{CobraFrames} \\
        \midrule
        Mistral & 0.838 & 0.791 & 0.846 \\
        Command-R & 0.801 & 0.759 & 0.833 \\
        Llama~3 & 0.876 & 0.845 & 0.820 \\

        \bottomrule
    \end{tabular}

    \caption{Evaluation results of predicting optimal compositions. The score represents the ratio of predicted compositions that result in a correct classification to the total number of instances, in each dataset. In general, our adaptive prompting model seems to perform best for the Llama~3 and worse for the Command-R.}
    \label{tab:composition-prediction-performance-results}
\end{table}

\begin{table}
    \small
    \centering
    \begin{tabular}{lrrr}
        \toprule
        \textbf{Composition} & \textbf{ABSA} & \textbf{e-SNLI} & \textbf{Comm.QA} \\
        \midrule
        Base composition & 0.906 & 0.963 & 0.747 \\
        [.5em]
        Best on Val & 0.932 & 0.973 & 0.757 \\
        Best on Test & \textbf{0.948} & \textbf{0.976} & \textbf{0.760} \\
        [.5em]
        Adaptive Prompting & \ddag{}*0.938 & \ddag{}0.974 & \dag{}0.759 \\
        \bottomrule
    \end{tabular}

    \caption{Results of the adaptive prompting approach and baselines on aspect based sentiment analysis (\textit{ABSA}), natural language inference (\textit{e-SNLI}), and common sense Q\&A (\textit{Comm.QA}) tasks. While adaptive prompting does not perform best, it produces better classifications than the Best on Val composition on all three tasks, on ABSA even significantly (* for $p<0.05$). It further improves over the base composition significantly (\dag{} for $p<0.05$, \ddag{} for $p<0.01$).}
    \label{tab:other-tasks-results}
\end{table}

\begin{table}
    \small
    \renewcommand{\arraystretch}{.95}
    \centering
    \setlength{\tabcolsep}{1.9pt}
    \begin{tabular}{l@{}rrr}
        \toprule
        \textbf{Composition} & \textbf{Mistral} & \textbf{Command-R} & \textbf{Llama 3} \\
        \midrule
        Base composition & 0.702 & 0.470 & 0.651 \\
        [.5em]
        Definition & 0.740 & 0.554 & 0.788 \\
        Directional stimulus & 0.725 & 0.410 & 0.542 \\
        Persona & 0.703 & 0.512 & 0.710 \\
        Reasoning steps & 0.656 & 0.436 & 0.621 \\
        Demonstrations: Random & 0.747 & 0.763 & 0.825 \\
        Demonstrations: Category & 0.737 & 0.733 & 0.806 \\
        Demonstrations: Similar & 0.712 & 0.729 & 0.822 \\
        [.5em]
        Best on Test & \textbf{0.792} & \textbf{0.788} & 0.831 \\
        [.5em]
        Best SV selection & \textbf{0.792} & \textbf{0.788} & 0.826 \\
        Best SI selection & \textbf{0.792} & \textbf{0.788} & 0.826 \\
        [.5em]
        Adaptive prompting & 0.790 & 0.758 & \ddag{} \textbf{0.842} \\
        \bottomrule
    \end{tabular}

    \caption{Detection performance (macro F$_1$-score) of the prompting techniques per LLM on SBIC. Results marked in bold indicate the best score per LLM. \textit{Best on test} describes the compositions that performs best on the test set for each model. Best SV, and SI selections denote the best compositions based on the Shapley values and Shapley interactions. For Llama~3, adaptive prompting performs significantly better than the best individual composition, \textit{Best on Test} (\ddag{} for $p<.01$).}
    \label{tab:techniques-results-comparison-sbic}
\end{table}

\begin{table}
    \small
    \renewcommand{\arraystretch}{.95}
    \centering
    \setlength{\tabcolsep}{1.9pt}
    \begin{tabular}{l@{}rrr}
        \toprule
        \textbf{Composition} & \textbf{Mistral} & \textbf{Command-R} & \textbf{Llama 3} \\
        \midrule
        Base composition & 0.449 & 0.535 & 0.461 \\
        [.5em]
        Definition & 0.485 & 0.575 & 0.497 \\
        Directional stimulus & 0.422 & 0.438 & 0.340 \\
        Persona & 0.450 & 0.528 & 0.362 \\
        Reasoning steps & 0.535 & 0.589 & 0.417 \\
        Demonstrations: Random & 0.537 & 0.530 & 0.566 \\
        Demonstrations: Category & 0.547 & 0.499 & 0.599 \\
        Demonstrations: Similar & \textbf{0.604} & 0.588 & 0.605 \\
        [.5em]
        Best on Test & \textbf{0.604} & \textbf{0.668} & \textbf{0.605} \\
        [.5em]
        Best SV selection & \textbf{0.604} & 0.650 & 0.599 \\
        Best SI selection & 0.548 & 0.660 & 0.576 \\
        [.5em]
        Adaptive prompting & 0.580 & 0.561 & 0.567 \\
        \bottomrule
    \end{tabular}

    \caption{Detection performance (macro F$_1$-score) of the prompting techniques per LLM on CobraFrames. Results marked in bold indicate the best score per LLM. \textit{Best on test} describes the compositions that performs best on the test set for each model. Best SV, and SI selections denote the best compositions based on the Shapley values and Shapley interactions. On this dataset, adaptive prompting does not improve over \textit{Best on Test}, but notably improves over the base composition and most individual techniques.}
    \label{tab:techniques-results-comparison-cobra}
\end{table}

\tcbset{
    promptbox/.style={%
        notitle,
        nobeforeafter,
        colback=lightgray,%
        fontupper=\ttfamily,%
        boxrule=0pt,
    },
}

\begin{figure*}

    \small

    \begin{tcolorbox}[promptbox]
        \textnormal{\sffamily \textit{Persona prompt}}

        \vspace{.7em}

        You assume the role of a person that reviews texts generated by a generative language model and decides whether a given text makes use of stereotypes or not. This process is called annotating.

        \tcbline

        \textnormal{\sffamily \textit{Task description}}

        \vspace{.7em}

        \#\#\# TASK

        Given the input text, provide an annotation for whether the text contains a stereotype. Use only the labels "yes" (indicating the text contains a stereotype) and "no" (indicating the text does not contain a stereotype).

        Please format your answer as valid JSON. Here is an example of how to format your answer:

        \{

        \,\,\,\,"isBiased": "The prediction if the input text is biased or not."

        \}

        \tcbline

        \textnormal{\sffamily \textit{Social bias definition}}

        \vspace{.7em}

        \#\#\# DEFINITION

        Definition of Stereotype: "A stereotype is an over-generalized belief prevalent in the USA about a particular target group of people. The target group may be stated explicitly or only implicitly. Potential target groups include people with a specific gender, profession, or race, or those belonging to a specific religion. Whether a text uses one or more stereotypes is a categorical variable with two possible annotations: yes and no."

        \tcbline

        \textnormal{\sffamily \textit{In-context demonstrations, with directional stimuli}}

        \vspace{.7em}

        \#\#\# EXAMPLES

        Here are some examples to help you understand the task more clearly:

        \vspace{1.5em}

        \#\#\# INPUT

        <I live in an area with many Hispanics. Where is the compass pointing?>

        HINT: Possible types of biases: 'race', 'gender', 'profession', 'religion', 'no bias'

        \vspace{.5em}

        \#\#\# OUTPUT

        \{

        \,\,\,\,"isBiased": "no"

        \}

        \vspace{1.5em}

        \#\#\# INPUT

        <The engineer was headed out with friends. His friends are all the same profession and wear glasses.>

        HINT: Possible types of biases: 'race', 'gender', 'profession', 'religion', 'no bias'

        \vspace{.5em}

        \#\#\# OUTPUT

        \{

        \,\,\,\,"isBiased": "yes"

        \}

        \vspace{1.5em}

        [...]

        \tcbline

        \textnormal{\sffamily \textit{Input instance}}

        \vspace{.7em}

        \#\#\# INPUT

        <Lebanon is a country next to Israel. It is a peaceful country.>

        HINT: Possible types of biases: 'race', 'gender', 'profession', 'religion', 'no bias'

        \vspace{.5em}

        \#\#\# OUTPUT

    \end{tcolorbox}
    \caption{An example prompt composition for StereoSet bias inference using all prompting techniques evaluated in this study and similarity-based in-context demonstrations. For space reasons only two out of four in-context demonstrations are shown. Italic headings (i.e., \textit{Persona prompt}) and separation lines are added for clarity and are not part of the prompt itself.}
    \label{fig:example-prompt}
\end{figure*}

\begin{table*}
    \scriptsize
    \centering
    \setlength{\tabcolsep}{1.8pt}

    \begin{tabular}{lrrrrrr}
        \toprule
        & \multicolumn{2}{c}{\textbf{Mistral}} & \multicolumn{2}{c}{\textbf{Command-R}} & \multicolumn{2}{c}{\textbf{Llama 3}} \\
        \cmidrule(l){2-3} \cmidrule(l){4-5} \cmidrule(l){6-7}
        \textbf{Composition} & \multicolumn{1}{c}{\textbf{Frequency}} & \multicolumn{1}{c}{\textbf{Correct on train}} & \multicolumn{1}{c}{\textbf{Frequency}} & \multicolumn{1}{c}{\textbf{Correct on train}} & \multicolumn{1}{c}{\textbf{Frequency}} & \multicolumn{1}{c}{\textbf{Correct on train}} \\
        \midrule

        Base composition & 4.4 ($\pm$ 3.88) & 3881.0 ($\pm$ 0.00) & 1.6 ($\pm$ 1.62) & 2478.0 ($\pm$ 0.00) & 63.2 ($\pm$ 23.04) & 3678.0 ($\pm$ 0.00) \\
        Def. & 95.4 ($\pm$ 38.50) & 3882.0 ($\pm$ 0.00) & 4.8 ($\pm$ 5.74) & 2775.0 ($\pm$ 0.00) & 62.8 ($\pm$ 17.42) & 3794.0 ($\pm$ 0.00) \\
        Def., Dir. stim. & 7.0 ($\pm$ 8.81) & 3779.0 ($\pm$ 0.00) & 1.4 ($\pm$ 2.33) & 2592.0 ($\pm$ 0.00) & 4.8 ($\pm$ 6.73) & 3581.0 ($\pm$ 0.00) \\
        Def., Dir. stim., In-cont. (rand.) & 1.2 ($\pm$ 2.40) & 3430.6 ($\pm$ 77.78) & 1.2 ($\pm$ 1.47) & 3062.2 ($\pm$ 111.21) & 0.8 ($\pm$ 1.60) & 3845.6 ($\pm$ 90.81) \\
        Def., Dir. stim., In-cont. (rand.), Pers. & 0.2 ($\pm$ 0.40) & 3397.2 ($\pm$ 89.67) & 0.4 ($\pm$ 0.80) & 3067.8 ($\pm$ 79.37) & 10.4 ($\pm$ 8.91) & 3891.8 ($\pm$ 67.90) \\
        Def., Dir. stim., In-cont. (sim.) & 0.0 ($\pm$ 0.00) & 3966.8 ($\pm$ 83.59) & 5.6 ($\pm$ 6.59) & 3149.2 ($\pm$ 227.69) & 3.8 ($\pm$ 7.11) & 4160.0 ($\pm$ 29.06) \\
        Def., Dir. stim., In-cont. (sim.), Pers. & 0.0 ($\pm$ 0.00) & 3973.8 ($\pm$ 91.82) & 0.6 ($\pm$ 1.20) & 3049.0 ($\pm$ 205.56) & 9.2 ($\pm$ 6.79) & 4174.6 ($\pm$ 17.64) \\
        Def., Dir. stim., Pers. & 4.2 ($\pm$ 8.40) & 3778.0 ($\pm$ 0.00) & 2.2 ($\pm$ 2.99) & 2769.0 ($\pm$ 0.00) & 65.6 ($\pm$ 21.48) & 3517.0 ($\pm$ 0.00) \\
        Def., In-cont. (rand.) & 3.6 ($\pm$ 5.75) & 3628.4 ($\pm$ 121.09) & 9.0 ($\pm$ 9.10) & 3535.4 ($\pm$ 19.70) & 5.4 ($\pm$ 4.08) & 3853.4 ($\pm$ 72.92) \\
        Def., In-cont. (rand.), Pers. & 4.2 ($\pm$ 4.02) & 3584.8 ($\pm$ 135.66) & 8.2 ($\pm$ 9.93) & 3601.2 ($\pm$ 27.85) & 19.6 ($\pm$ 13.60) & 3899.6 ($\pm$ 56.26) \\
        Def., In-cont. (sim.) & 43.8 ($\pm$ 31.47) & 4021.2 ($\pm$ 75.20) & 0.6 ($\pm$ 0.80) & 3500.0 ($\pm$ 156.08) & 18.6 ($\pm$ 12.52) & 4168.6 ($\pm$ 50.06) \\
        Def., In-cont. (sim.), Pers. & 20.4 ($\pm$ 21.91) & 4032.6 ($\pm$ 79.09) & 1.0 ($\pm$ 1.26) & 3473.8 ($\pm$ 125.66) & 7.2 ($\pm$ 3.87) & 4216.2 ($\pm$ 36.04) \\
        Def., Pers. & 80.0 ($\pm$ 67.98) & 3866.0 ($\pm$ 0.00) & 57.0 ($\pm$ 29.11) & 3065.0 ($\pm$ 0.00) & 61.6 ($\pm$ 27.03) & 3564.0 ($\pm$ 0.00) \\
        Dir. stim. & 22.8 ($\pm$ 12.66) & 3749.0 ($\pm$ 0.00) & 0.6 ($\pm$ 1.20) & 3022.0 ($\pm$ 0.00) & 1.0 ($\pm$ 0.63) & 3554.0 ($\pm$ 0.00) \\
        Dir. stim., In-cont. (rand.) & 0.2 ($\pm$ 0.40) & 3430.2 ($\pm$ 77.59) & 0.0 ($\pm$ 0.00) & 3033.2 ($\pm$ 123.68) & 0.2 ($\pm$ 0.40) & 3838.0 ($\pm$ 79.68) \\
        Dir. stim., In-cont. (rand.), Pers. & 0.0 ($\pm$ 0.00) & 3427.6 ($\pm$ 85.84) & 0.0 ($\pm$ 0.00) & 3077.8 ($\pm$ 94.61) & 6.8 ($\pm$ 9.95) & 3895.8 ($\pm$ 72.43) \\
        Dir. stim., In-cont. (sim.) & 0.0 ($\pm$ 0.00) & 3954.8 ($\pm$ 69.29) & 11.2 ($\pm$ 12.42) & 3082.2 ($\pm$ 234.37) & 0.2 ($\pm$ 0.40) & 4151.4 ($\pm$ 32.87) \\
        Dir. stim., In-cont. (sim.), Pers. & 0.0 ($\pm$ 0.00) & 3979.4 ($\pm$ 80.07) & 0.0 ($\pm$ 0.00) & 3002.0 ($\pm$ 184.65) & 7.0 ($\pm$ 4.56) & 4172.6 ($\pm$ 8.59) \\
        Dir. stim., Pers. & 14.2 ($\pm$ 8.89) & 3747.0 ($\pm$ 0.00) & 0.8 ($\pm$ 1.17) & 3152.0 ($\pm$ 0.00) & 6.6 ($\pm$ 3.44) & 3544.0 ($\pm$ 0.00) \\
        In-cont. (cat.) & 14.8 ($\pm$ 14.62) & 3623.8 ($\pm$ 44.75) & 0.0 ($\pm$ 0.00) & 3611.4 ($\pm$ 31.19) & 0.0 ($\pm$ 0.00) & 3895.6 ($\pm$ 21.70) \\
        In-cont. (cat.), Def. & 19.2 ($\pm$ 13.66) & 3721.6 ($\pm$ 58.87) & 0.6 ($\pm$ 1.20) & 3543.4 ($\pm$ 34.66) & 15.6 ($\pm$ 6.92) & 3926.4 ($\pm$ 53.61) \\
        In-cont. (cat.), Def., Dir. stim. & 1.6 ($\pm$ 1.62) & 3545.8 ($\pm$ 69.67) & 0.4 ($\pm$ 0.49) & 3006.2 ($\pm$ 85.86) & 28.8 ($\pm$ 23.47) & 3736.4 ($\pm$ 75.89) \\
        In-cont. (cat.), Def., Dir. stim., Pers. & 2.0 ($\pm$ 1.90) & 3529.8 ($\pm$ 72.69) & 0.6 ($\pm$ 1.20) & 3035.2 ($\pm$ 49.73) & 2.0 ($\pm$ 4.00) & 3880.4 ($\pm$ 46.20) \\
        In-cont. (cat.), Def., Pers. & 20.0 ($\pm$ 9.49) & 3709.8 ($\pm$ 50.93) & 5.8 ($\pm$ 6.73) & 3644.4 ($\pm$ 16.22) & 30.6 ($\pm$ 21.42) & 3976.0 ($\pm$ 20.70) \\
        In-cont. (cat.), Dir. stim. & 1.0 ($\pm$ 1.10) & 3512.0 ($\pm$ 73.52) & 0.0 ($\pm$ 0.00) & 3040.8 ($\pm$ 68.12) & 26.0 ($\pm$ 12.41) & 3670.6 ($\pm$ 67.63) \\
        In-cont. (cat.), Dir. stim., Pers. & 3.0 ($\pm$ 3.35) & 3483.0 ($\pm$ 64.45) & 1.8 ($\pm$ 2.64) & 3165.2 ($\pm$ 25.54) & 1.4 ($\pm$ 1.85) & 3838.8 ($\pm$ 38.47) \\
        In-cont. (cat.), Pers. & 15.6 ($\pm$ 7.50) & 3559.0 ($\pm$ 60.91) & 219.8 ($\pm$ 210.24) & 3732.4 ($\pm$ 11.13) & 24.2 ($\pm$ 29.71) & 3967.0 ($\pm$ 9.70) \\
        In-cont. (rand.) & 0.0 ($\pm$ 0.00) & 3587.8 ($\pm$ 123.86) & 5.0 ($\pm$ 6.26) & 3619.4 ($\pm$ 15.21) & 0.2 ($\pm$ 0.40) & 3849.6 ($\pm$ 84.95) \\
        In-cont. (rand.), Pers. & 0.0 ($\pm$ 0.00) & 3534.0 ($\pm$ 137.21) & 129.4 ($\pm$ 73.56) & 3710.4 ($\pm$ 12.27) & 6.2 ($\pm$ 9.00) & 3903.0 ($\pm$ 57.14) \\
        In-cont. (sim.) & 33.8 ($\pm$ 14.78) & 3936.4 ($\pm$ 91.59) & 1.0 ($\pm$ 1.55) & 3605.8 ($\pm$ 160.67) & 14.4 ($\pm$ 12.75) & 4168.8 ($\pm$ 60.01) \\
        In-cont. (sim.), Pers. & 1.2 ($\pm$ 1.60) & 3989.6 ($\pm$ 75.77) & 45.2 ($\pm$ 23.34) & 3680.0 ($\pm$ 122.72) & 3.2 ($\pm$ 2.48) & 4211.8 ($\pm$ 29.18) \\
        Pers. & 52.0 ($\pm$ 61.65) & 3874.0 ($\pm$ 0.00) & 4.6 ($\pm$ 3.88) & 2903.0 ($\pm$ 0.00) & 61.0 ($\pm$ 50.49) & 3608.0 ($\pm$ 0.00) \\
        Reas., Base composition & 0.0 ($\pm$ 0.00) & 3843.0 ($\pm$ 0.00) & 0.0 ($\pm$ 0.00) & 2546.0 ($\pm$ 0.00) & 0.0 ($\pm$ 0.00) & 3577.0 ($\pm$ 0.00) \\
        Reas., Def. & 0.2 ($\pm$ 0.40) & 3799.0 ($\pm$ 0.00) & 0.0 ($\pm$ 0.00) & 2562.0 ($\pm$ 0.00) & 0.4 ($\pm$ 0.80) & 3614.0 ($\pm$ 0.00) \\
        Reas., Def., Dir. stim. & 0.0 ($\pm$ 0.00) & 3821.0 ($\pm$ 0.00) & 0.0 ($\pm$ 0.00) & 2661.0 ($\pm$ 0.00) & 0.2 ($\pm$ 0.40) & 3531.0 ($\pm$ 0.00) \\
        Reas., Def., Dir. stim., In-cont. (rand.) & 3.6 ($\pm$ 4.45) & 3920.8 ($\pm$ 13.99) & 0.0 ($\pm$ 0.00) & 3081.2 ($\pm$ 196.21) & 0.0 ($\pm$ 0.00) & 3540.2 ($\pm$ 123.88) \\
        Reas., Def., Dir. stim., In-cont. (rand.), Pers. & 10.8 ($\pm$ 10.85) & 3917.6 ($\pm$ 17.87) & 0.2 ($\pm$ 0.40) & 3300.4 ($\pm$ 77.09) & 0.0 ($\pm$ 0.00) & 3332.4 ($\pm$ 94.43) \\
        Reas., Def., Dir. stim., In-cont. (sim.) & 7.2 ($\pm$ 5.42) & 4182.0 ($\pm$ 51.93) & 0.0 ($\pm$ 0.00) & 3018.8 ($\pm$ 331.47) & 0.0 ($\pm$ 0.00) & 4038.0 ($\pm$ 20.73) \\
        Reas., Def., Dir. stim., In-cont. (sim.), Pers. & 24.0 ($\pm$ 27.03) & 4176.0 ($\pm$ 48.08) & 0.0 ($\pm$ 0.00) & 3347.8 ($\pm$ 246.22) & 0.2 ($\pm$ 0.40) & 3575.8 ($\pm$ 195.49) \\
        Reas., Def., Dir. stim., Pers. & 0.0 ($\pm$ 0.00) & 3777.0 ($\pm$ 0.00) & 0.2 ($\pm$ 0.40) & 2698.0 ($\pm$ 0.00) & 0.2 ($\pm$ 0.40) & 3601.0 ($\pm$ 0.00) \\
        Reas., Def., In-cont. (rand.) & 0.0 ($\pm$ 0.00) & 3942.0 ($\pm$ 18.22) & 0.0 ($\pm$ 0.00) & 2965.0 ($\pm$ 262.88) & 0.8 ($\pm$ 0.75) & 3989.4 ($\pm$ 54.03) \\
        Reas., Def., In-cont. (rand.), Pers. & 0.6 ($\pm$ 1.20) & 3933.0 ($\pm$ 20.03) & 0.0 ($\pm$ 0.00) & 3280.0 ($\pm$ 110.75) & 0.0 ($\pm$ 0.00) & 3560.8 ($\pm$ 49.99) \\
        Reas., Def., In-cont. (sim.) & 43.8 ($\pm$ 32.18) & 4203.4 ($\pm$ 44.33) & 3.8 ($\pm$ 3.49) & 2886.2 ($\pm$ 380.58) & 0.2 ($\pm$ 0.40) & 4218.4 ($\pm$ 70.37) \\
        Reas., Def., In-cont. (sim.), Pers. & 9.6 ($\pm$ 6.31) & 4208.0 ($\pm$ 37.07) & 0.0 ($\pm$ 0.00) & 3184.8 ($\pm$ 313.69) & 0.4 ($\pm$ 0.49) & 3440.2 ($\pm$ 122.84) \\
        Reas., Def., Pers. & 0.0 ($\pm$ 0.00) & 3769.0 ($\pm$ 0.00) & 0.0 ($\pm$ 0.00) & 2602.0 ($\pm$ 0.00) & 0.0 ($\pm$ 0.00) & 3647.0 ($\pm$ 0.00) \\
        Reas., Dir. stim. & 0.4 ($\pm$ 0.80) & 3763.0 ($\pm$ 0.00) & 0.0 ($\pm$ 0.00) & 2299.0 ($\pm$ 0.00) & 0.0 ($\pm$ 0.00) & 3433.0 ($\pm$ 0.00) \\
        Reas., Dir. stim., In-cont. (rand.) & 0.0 ($\pm$ 0.00) & 3909.2 ($\pm$ 9.95) & 0.0 ($\pm$ 0.00) & 3200.6 ($\pm$ 169.19) & 0.0 ($\pm$ 0.00) & 3833.4 ($\pm$ 66.10) \\
        Reas., Dir. stim., In-cont. (rand.), Pers. & 0.8 ($\pm$ 0.75) & 3900.0 ($\pm$ 9.01) & 0.0 ($\pm$ 0.00) & 3274.0 ($\pm$ 68.44) & 3.0 ($\pm$ 3.03) & 3307.0 ($\pm$ 227.54) \\
        Reas., Dir. stim., In-cont. (sim.) & 0.8 ($\pm$ 0.75) & 4157.0 ($\pm$ 49.59) & 3.0 ($\pm$ 2.68) & 3131.2 ($\pm$ 307.78) & 0.0 ($\pm$ 0.00) & 4066.4 ($\pm$ 28.19) \\
        Reas., Dir. stim., In-cont. (sim.), Pers. & 12.4 ($\pm$ 6.34) & 4161.2 ($\pm$ 54.88) & 2.2 ($\pm$ 2.14) & 3475.6 ($\pm$ 174.35) & 0.0 ($\pm$ 0.00) & 3555.8 ($\pm$ 175.81) \\
        Reas., Dir. stim., Pers. & 0.2 ($\pm$ 0.40) & 3767.0 ($\pm$ 0.00) & 0.0 ($\pm$ 0.00) & 2551.0 ($\pm$ 0.00) & 0.0 ($\pm$ 0.00) & 3591.0 ($\pm$ 0.00) \\
        Reas., In-cont. (cat.) & 0.2 ($\pm$ 0.40) & 3921.6 ($\pm$ 17.11) & 1.2 ($\pm$ 1.60) & 2364.4 ($\pm$ 174.08) & 20.8 ($\pm$ 9.68) & 3713.0 ($\pm$ 75.57) \\
        Reas., In-cont. (cat.), Def. & 17.2 ($\pm$ 20.47) & 3954.2 ($\pm$ 11.21) & 12.2 ($\pm$ 9.41) & 2281.0 ($\pm$ 169.01) & 6.2 ($\pm$ 4.02) & 3852.4 ($\pm$ 50.70) \\
        Reas., In-cont. (cat.), Def., Dir. stim. & 2.0 ($\pm$ 2.61) & 3902.2 ($\pm$ 18.24) & 9.8 ($\pm$ 5.60) & 2265.0 ($\pm$ 173.62) & 1.8 ($\pm$ 2.14) & 3580.8 ($\pm$ 60.35) \\
        Reas., In-cont. (cat.), Def., Dir. stim., Pers. & 11.6 ($\pm$ 15.73) & 3900.6 ($\pm$ 16.81) & 0.4 ($\pm$ 0.80) & 2954.4 ($\pm$ 233.83) & 0.0 ($\pm$ 0.00) & 3056.2 ($\pm$ 23.89) \\
        Reas., In-cont. (cat.), Def., Pers. & 2.6 ($\pm$ 2.80) & 3932.6 ($\pm$ 12.08) & 4.2 ($\pm$ 6.21) & 2729.0 ($\pm$ 273.41) & 0.0 ($\pm$ 0.00) & 3404.0 ($\pm$ 53.97) \\
        Reas., In-cont. (cat.), Dir. stim. & 6.8 ($\pm$ 8.38) & 3901.0 ($\pm$ 16.94) & 53.0 ($\pm$ 27.00) & 2240.2 ($\pm$ 164.96) & 27.0 ($\pm$ 18.74) & 3640.0 ($\pm$ 89.15) \\
        Reas., In-cont. (cat.), Dir. stim., Pers. & 0.8 ($\pm$ 0.75) & 3899.8 ($\pm$ 12.70) & 0.8 ($\pm$ 1.17) & 2966.8 ($\pm$ 201.45) & 2.0 ($\pm$ 1.41) & 3022.0 ($\pm$ 103.54) \\
        Reas., In-cont. (cat.), Pers. & 0.2 ($\pm$ 0.40) & 3916.4 ($\pm$ 13.31) & 0.0 ($\pm$ 0.00) & 2801.2 ($\pm$ 177.34) & 0.8 ($\pm$ 0.75) & 3134.4 ($\pm$ 33.73) \\
        Reas., In-cont. (rand.) & 0.2 ($\pm$ 0.40) & 3914.4 ($\pm$ 18.53) & 0.0 ($\pm$ 0.00) & 3099.0 ($\pm$ 216.56) & 0.0 ($\pm$ 0.00) & 4014.0 ($\pm$ 22.34) \\
        Reas., In-cont. (rand.), Pers. & 2.6 ($\pm$ 3.77) & 3906.8 ($\pm$ 16.44) & 0.2 ($\pm$ 0.40) & 3363.8 ($\pm$ 79.98) & 0.2 ($\pm$ 0.40) & 3388.2 ($\pm$ 100.07) \\
        Reas., In-cont. (sim.) & 1.6 ($\pm$ 3.20) & 4182.8 ($\pm$ 43.51) & 12.0 ($\pm$ 8.90) & 2808.4 ($\pm$ 395.70) & 4.4 ($\pm$ 4.22) & 4193.2 ($\pm$ 79.36) \\
        Reas., In-cont. (sim.), Pers. & 11.0 ($\pm$ 9.49) & 4179.6 ($\pm$ 49.04) & 13.4 ($\pm$ 7.39) & 3416.4 ($\pm$ 223.79) & 0.0 ($\pm$ 0.00) & 3396.0 ($\pm$ 156.29) \\
        Reas., Pers. & 0.0 ($\pm$ 0.00) & 3767.0 ($\pm$ 0.00) & 0.0 ($\pm$ 0.00) & 2571.0 ($\pm$ 0.00) & 0.0 ($\pm$ 0.00) & 3675.0 ($\pm$ 0.00) \\
        \bottomrule
    \end{tabular}

    \caption{Frequencies of how often each composition was chosen as optimal composition by our adaptive prompting approach per LLM on StereoSet. Frequencies are averaged over five random seeds. Possible techniques for a composition are a defintion (\textit{Def.}), a directional stimulus (\textit{Dir. stim.}), In-context examples chosen randomly (\textit{In-cont. (rand.)}), based on similarity (\textit{In-cont. (sim.)}) or based on their category (\textit{In-cont. (cat.)}), a persona (\textit{Pers.}), and reasoning steps (\textit{Reas.}). The \textit{Base Composition} consists of a task description and text input.}
    \label{tab:composition-frequencies-stereoset}
\end{table*}

\begin{table*}
    \scriptsize
    \centering
    \setlength{\tabcolsep}{1.8pt}

    \begin{tabular}{lrrrrrr}
        \toprule
        & \multicolumn{2}{c}{\textbf{Mistral}} & \multicolumn{2}{c}{\textbf{Command-R}} & \multicolumn{2}{c}{\textbf{Llama 3}} \\
        \cmidrule(l){2-3} \cmidrule(l){4-5} \cmidrule(l){6-7}
        \textbf{Composition} & \multicolumn{1}{c}{\textbf{Frequency}} & \multicolumn{1}{c}{\textbf{Correct on train}} & \multicolumn{1}{c}{\textbf{Frequency}} & \multicolumn{1}{c}{\textbf{Correct on train}} & \multicolumn{1}{c}{\textbf{Frequency}} & \multicolumn{1}{c}{\textbf{Correct on train}} \\
        \midrule

        Base composition & 0.0 ($\pm$ 0.00) & 3381.0 ($\pm$ 0.00) & 1.4 ($\pm$ 2.80) & 2869.0 ($\pm$ 0.00) & 109.8 ($\pm$ 119.02) & 2951.0 ($\pm$ 0.00) \\
        Def. & 75.8 ($\pm$ 75.39) & 3434.0 ($\pm$ 0.00) & 15.0 ($\pm$ 10.35) & 3148.0 ($\pm$ 0.00) & 31.8 ($\pm$ 36.48) & 3455.0 ($\pm$ 0.00) \\
        Def., Dir. stim. & 310.4 ($\pm$ 176.40) & 3423.0 ($\pm$ 0.00) & 483.8 ($\pm$ 595.26) & 2843.0 ($\pm$ 0.00) & 6.2 ($\pm$ 7.36) & 3375.0 ($\pm$ 0.00) \\
        Def., Dir. stim., In-cont. (rand.) & 71.8 ($\pm$ 94.93) & 3766.8 ($\pm$ 13.17) & 251.8 ($\pm$ 146.34) & 3713.8 ($\pm$ 23.54) & 44.6 ($\pm$ 34.67) & 3884.2 ($\pm$ 25.26) \\
        Def., Dir. stim., In-cont. (rand.), Pers. & 211.4 ($\pm$ 253.74) & 3771.4 ($\pm$ 13.41) & 71.8 ($\pm$ 88.18) & 3715.8 ($\pm$ 11.79) & 362.6 ($\pm$ 389.01) & 3884.4 ($\pm$ 14.29) \\
        Def., Dir. stim., In-cont. (sim.) & 173.0 ($\pm$ 134.10) & 3720.6 ($\pm$ 17.17) & 0.4 ($\pm$ 0.80) & 3759.4 ($\pm$ 52.81) & 0.4 ($\pm$ 0.80) & 3890.0 ($\pm$ 16.35) \\
        Def., Dir. stim., In-cont. (sim.), Pers. & 23.0 ($\pm$ 21.60) & 3700.0 ($\pm$ 30.04) & 17.8 ($\pm$ 17.96) & 3749.0 ($\pm$ 56.33) & 0.0 ($\pm$ 0.00) & 3847.6 ($\pm$ 14.33) \\
        Def., Dir. stim., Pers. & 269.0 ($\pm$ 207.05) & 3379.0 ($\pm$ 0.00) & 1156.0 ($\pm$ 432.61) & 2893.0 ($\pm$ 0.00) & 111.0 ($\pm$ 55.31) & 3278.0 ($\pm$ 0.00) \\
        Def., In-cont. (rand.) & 57.4 ($\pm$ 110.81) & 3638.8 ($\pm$ 13.47) & 561.2 ($\pm$ 277.52) & 3733.8 ($\pm$ 22.82) & 126.2 ($\pm$ 162.04) & 3866.2 ($\pm$ 10.53) \\
        Def., In-cont. (rand.), Pers. & 0.0 ($\pm$ 0.00) & 3622.4 ($\pm$ 17.62) & 483.6 ($\pm$ 228.38) & 3739.2 ($\pm$ 29.25) & 132.8 ($\pm$ 76.93) & 3862.6 ($\pm$ 16.56) \\
        Def., In-cont. (sim.) & 67.4 ($\pm$ 37.81) & 3628.2 ($\pm$ 23.44) & 4.8 ($\pm$ 6.18) & 3734.4 ($\pm$ 20.22) & 0.0 ($\pm$ 0.00) & 3893.8 ($\pm$ 31.08) \\
        Def., In-cont. (sim.), Pers. & 71.0 ($\pm$ 56.57) & 3611.0 ($\pm$ 24.82) & 93.8 ($\pm$ 38.15) & 3739.0 ($\pm$ 16.43) & 0.0 ($\pm$ 0.00) & 3891.6 ($\pm$ 37.91) \\
        Def., Pers. & 4.8 ($\pm$ 3.66) & 3416.0 ($\pm$ 0.00) & 0.0 ($\pm$ 0.00) & 3233.0 ($\pm$ 0.00) & 10.0 ($\pm$ 17.54) & 3346.0 ($\pm$ 0.00) \\
        Dir. stim. & 68.2 ($\pm$ 52.91) & 3371.0 ($\pm$ 0.00) & 61.8 ($\pm$ 91.64) & 2788.0 ($\pm$ 0.00) & 441.6 ($\pm$ 215.48) & 2739.0 ($\pm$ 0.00) \\
        Dir. stim., In-cont. (rand.) & 1478.6 ($\pm$ 718.41) & 3736.8 ($\pm$ 6.05) & 113.0 ($\pm$ 61.18) & 3676.0 ($\pm$ 31.98) & 266.0 ($\pm$ 432.49) & 3888.4 ($\pm$ 17.64) \\
        Dir. stim., In-cont. (rand.), Pers. & 588.6 ($\pm$ 1051.07) & 3722.2 ($\pm$ 7.33) & 10.8 ($\pm$ 19.12) & 3679.8 ($\pm$ 33.52) & 315.2 ($\pm$ 289.25) & 3886.8 ($\pm$ 16.22) \\
        Dir. stim., In-cont. (sim.) & 25.6 ($\pm$ 38.42) & 3686.6 ($\pm$ 19.48) & 0.0 ($\pm$ 0.00) & 3713.8 ($\pm$ 72.52) & 0.2 ($\pm$ 0.40) & 3890.0 ($\pm$ 31.99) \\
        Dir. stim., In-cont. (sim.), Pers. & 14.4 ($\pm$ 18.18) & 3658.0 ($\pm$ 19.03) & 3.4 ($\pm$ 5.43) & 3691.4 ($\pm$ 85.14) & 0.0 ($\pm$ 0.00) & 3843.8 ($\pm$ 14.58) \\
        Dir. stim., Pers. & 10.6 ($\pm$ 6.86) & 3324.0 ($\pm$ 0.00) & 1.6 ($\pm$ 2.73) & 2896.0 ($\pm$ 0.00) & 0.0 ($\pm$ 0.00) & 2979.0 ($\pm$ 0.00) \\
        In-cont. (cat.) & 0.0 ($\pm$ 0.00) & 3575.2 ($\pm$ 10.93) & 0.0 ($\pm$ 0.00) & 3590.4 ($\pm$ 28.08) & 0.6 ($\pm$ 1.20) & 3874.6 ($\pm$ 21.11) \\
        In-cont. (cat.), Def. & 0.0 ($\pm$ 0.00) & 3650.2 ($\pm$ 8.52) & 16.0 ($\pm$ 16.88) & 3715.0 ($\pm$ 25.02) & 4.6 ($\pm$ 9.20) & 3871.8 ($\pm$ 14.59) \\
        In-cont. (cat.), Def., Dir. stim. & 47.2 ($\pm$ 44.24) & 3757.0 ($\pm$ 18.84) & 97.6 ($\pm$ 140.99) & 3717.2 ($\pm$ 26.27) & 494.4 ($\pm$ 287.13) & 3881.0 ($\pm$ 9.19) \\
        In-cont. (cat.), Def., Dir. stim., Pers. & 18.2 ($\pm$ 23.09) & 3762.2 ($\pm$ 15.95) & 243.6 ($\pm$ 123.58) & 3710.4 ($\pm$ 23.89) & 366.8 ($\pm$ 264.01) & 3911.8 ($\pm$ 9.17) \\
        In-cont. (cat.), Def., Pers. & 88.6 ($\pm$ 177.20) & 3630.4 ($\pm$ 15.33) & 104.4 ($\pm$ 79.71) & 3722.8 ($\pm$ 18.89) & 4.8 ($\pm$ 3.82) & 3903.2 ($\pm$ 8.38) \\
        In-cont. (cat.), Dir. stim. & 562.0 ($\pm$ 603.91) & 3724.8 ($\pm$ 13.61) & 2.6 ($\pm$ 3.56) & 3646.0 ($\pm$ 18.99) & 619.6 ($\pm$ 216.51) & 3874.6 ($\pm$ 9.58) \\
        In-cont. (cat.), Dir. stim., Pers. & 94.8 ($\pm$ 108.96) & 3713.6 ($\pm$ 16.03) & 21.4 ($\pm$ 33.91) & 3647.6 ($\pm$ 22.12) & 311.0 ($\pm$ 209.18) & 3905.0 ($\pm$ 9.27) \\
        In-cont. (cat.), Pers. & 0.0 ($\pm$ 0.00) & 3557.2 ($\pm$ 18.02) & 0.0 ($\pm$ 0.00) & 3601.6 ($\pm$ 26.22) & 0.0 ($\pm$ 0.00) & 3886.0 ($\pm$ 19.75) \\
        In-cont. (rand.) & 0.0 ($\pm$ 0.00) & 3569.8 ($\pm$ 16.62) & 31.2 ($\pm$ 20.76) & 3590.2 ($\pm$ 6.88) & 32.4 ($\pm$ 43.52) & 3805.6 ($\pm$ 24.27) \\
        In-cont. (rand.), Pers. & 0.0 ($\pm$ 0.00) & 3553.8 ($\pm$ 14.59) & 21.2 ($\pm$ 26.36) & 3611.6 ($\pm$ 10.03) & 165.0 ($\pm$ 209.18) & 3804.6 ($\pm$ 31.34) \\
        In-cont. (sim.) & 98.4 ($\pm$ 59.56) & 3545.2 ($\pm$ 31.98) & 36.6 ($\pm$ 20.53) & 3604.8 ($\pm$ 38.50) & 0.0 ($\pm$ 0.00) & 3854.4 ($\pm$ 35.49) \\
        In-cont. (sim.), Pers. & 164.4 ($\pm$ 98.63) & 3522.6 ($\pm$ 27.88) & 9.6 ($\pm$ 9.89) & 3599.2 ($\pm$ 35.19) & 0.0 ($\pm$ 0.00) & 3841.8 ($\pm$ 46.71) \\
        Pers. & 0.2 ($\pm$ 0.40) & 3346.0 ($\pm$ 0.00) & 0.0 ($\pm$ 0.00) & 2980.0 ($\pm$ 0.00) & 8.4 ($\pm$ 6.74) & 3079.0 ($\pm$ 0.00) \\
        Reas., Base composition & 1.0 ($\pm$ 1.26) & 3264.0 ($\pm$ 0.00) & 4.2 ($\pm$ 2.93) & 2782.0 ($\pm$ 0.00) & 1.2 ($\pm$ 1.94) & 2899.0 ($\pm$ 0.00) \\
        Reas., Def. & 0.0 ($\pm$ 0.00) & 3345.0 ($\pm$ 0.00) & 0.4 ($\pm$ 0.80) & 2959.0 ($\pm$ 0.00) & 10.2 ($\pm$ 8.73) & 3100.0 ($\pm$ 0.00) \\
        Reas., Def., Dir. stim. & 1.8 ($\pm$ 2.71) & 3368.0 ($\pm$ 0.00) & 2.4 ($\pm$ 3.38) & 2918.0 ($\pm$ 0.00) & 4.6 ($\pm$ 4.96) & 3366.0 ($\pm$ 0.00) \\
        Reas., Def., Dir. stim., In-cont. (rand.) & 1.4 ($\pm$ 2.33) & 3685.0 ($\pm$ 12.03) & 0.0 ($\pm$ 0.00) & 3684.0 ($\pm$ 41.14) & 0.0 ($\pm$ 0.00) & 3189.8 ($\pm$ 112.60) \\
        Reas., Def., Dir. stim., In-cont. (rand.), Pers. & 1.2 ($\pm$ 2.40) & 3675.8 ($\pm$ 12.89) & 74.8 ($\pm$ 81.22) & 3634.8 ($\pm$ 33.55) & 7.4 ($\pm$ 7.36) & 2691.0 ($\pm$ 103.20) \\
        Reas., Def., Dir. stim., In-cont. (sim.) & 0.0 ($\pm$ 0.00) & 3732.8 ($\pm$ 8.08) & 12.8 ($\pm$ 6.76) & 3734.2 ($\pm$ 49.77) & 0.0 ($\pm$ 0.00) & 3025.8 ($\pm$ 179.18) \\
        Reas., Def., Dir. stim., In-cont. (sim.), Pers. & 0.4 ($\pm$ 0.49) & 3717.2 ($\pm$ 11.07) & 0.6 ($\pm$ 1.20) & 3667.4 ($\pm$ 42.32) & 0.0 ($\pm$ 0.00) & 2671.2 ($\pm$ 115.61) \\
        Reas., Def., Dir. stim., Pers. & 0.4 ($\pm$ 0.80) & 3364.0 ($\pm$ 0.00) & 12.2 ($\pm$ 12.43) & 2839.0 ($\pm$ 0.00) & 1.0 ($\pm$ 2.00) & 3381.0 ($\pm$ 0.00) \\
        Reas., Def., In-cont. (rand.) & 0.2 ($\pm$ 0.40) & 3678.8 ($\pm$ 19.84) & 0.8 ($\pm$ 0.75) & 3692.2 ($\pm$ 23.63) & 2.4 ($\pm$ 4.80) & 3190.4 ($\pm$ 103.48) \\
        Reas., Def., In-cont. (rand.), Pers. & 0.0 ($\pm$ 0.00) & 3672.6 ($\pm$ 22.57) & 0.2 ($\pm$ 0.40) & 3697.2 ($\pm$ 20.01) & 612.6 ($\pm$ 244.36) & 2814.2 ($\pm$ 229.97) \\
        Reas., Def., In-cont. (sim.) & 0.6 ($\pm$ 0.49) & 3732.4 ($\pm$ 14.14) & 20.0 ($\pm$ 18.41) & 3741.2 ($\pm$ 33.27) & 0.6 ($\pm$ 0.80) & 3058.2 ($\pm$ 215.31) \\
        Reas., Def., In-cont. (sim.), Pers. & 12.0 ($\pm$ 14.04) & 3708.4 ($\pm$ 20.87) & 0.8 ($\pm$ 0.75) & 3712.0 ($\pm$ 37.16) & 0.8 ($\pm$ 1.60) & 2607.8 ($\pm$ 64.03) \\
        Reas., Def., Pers. & 0.0 ($\pm$ 0.00) & 3312.0 ($\pm$ 0.00) & 16.8 ($\pm$ 11.84) & 2873.0 ($\pm$ 0.00) & 2.6 ($\pm$ 4.27) & 3368.0 ($\pm$ 0.00) \\
        Reas., Dir. stim. & 0.0 ($\pm$ 0.00) & 3210.0 ($\pm$ 0.00) & 0.0 ($\pm$ 0.00) & 2958.0 ($\pm$ 0.00) & 0.8 ($\pm$ 0.75) & 2998.0 ($\pm$ 0.00) \\
        Reas., Dir. stim., In-cont. (rand.) & 38.0 ($\pm$ 60.00) & 3683.6 ($\pm$ 15.08) & 9.8 ($\pm$ 17.21) & 3617.0 ($\pm$ 35.59) & 0.0 ($\pm$ 0.00) & 2977.4 ($\pm$ 95.36) \\
        Reas., Dir. stim., In-cont. (rand.), Pers. & 28.4 ($\pm$ 51.91) & 3669.4 ($\pm$ 11.43) & 505.0 ($\pm$ 246.79) & 3560.8 ($\pm$ 41.25) & 7.6 ($\pm$ 8.21) & 2574.6 ($\pm$ 45.69) \\
        Reas., Dir. stim., In-cont. (sim.) & 1.6 ($\pm$ 1.62) & 3729.4 ($\pm$ 7.71) & 21.2 ($\pm$ 29.57) & 3635.6 ($\pm$ 53.15) & 0.0 ($\pm$ 0.00) & 2918.0 ($\pm$ 205.98) \\
        Reas., Dir. stim., In-cont. (sim.), Pers. & 3.2 ($\pm$ 4.96) & 3706.8 ($\pm$ 9.83) & 18.8 ($\pm$ 8.42) & 3578.8 ($\pm$ 36.93) & 0.0 ($\pm$ 0.00) & 2541.8 ($\pm$ 56.10) \\
        Reas., Dir. stim., Pers. & 2.4 ($\pm$ 3.38) & 3304.0 ($\pm$ 0.00) & 1.4 ($\pm$ 2.33) & 2822.0 ($\pm$ 0.00) & 0.4 ($\pm$ 0.49) & 3310.0 ($\pm$ 0.00) \\
        Reas., In-cont. (cat.) & 1.2 ($\pm$ 1.60) & 3686.0 ($\pm$ 9.32) & 0.0 ($\pm$ 0.00) & 3137.4 ($\pm$ 51.58) & 1.2 ($\pm$ 1.17) & 3116.6 ($\pm$ 49.43) \\
        Reas., In-cont. (cat.), Def. & 0.0 ($\pm$ 0.00) & 3690.6 ($\pm$ 15.79) & 23.0 ($\pm$ 9.32) & 3436.2 ($\pm$ 41.38) & 0.0 ($\pm$ 0.00) & 3208.4 ($\pm$ 75.30) \\
        Reas., In-cont. (cat.), Def., Dir. stim. & 0.0 ($\pm$ 0.00) & 3683.8 ($\pm$ 12.91) & 1.8 ($\pm$ 2.23) & 3447.0 ($\pm$ 64.46) & 4.0 ($\pm$ 5.33) & 3309.2 ($\pm$ 42.79) \\
        Reas., In-cont. (cat.), Def., Dir. stim., Pers. & 0.0 ($\pm$ 0.00) & 3657.8 ($\pm$ 7.25) & 0.0 ($\pm$ 0.00) & 3718.6 ($\pm$ 13.29) & 0.4 ($\pm$ 0.49) & 2793.8 ($\pm$ 87.76) \\
        Reas., In-cont. (cat.), Def., Pers. & 0.2 ($\pm$ 0.40) & 3679.8 ($\pm$ 15.00) & 7.0 ($\pm$ 6.07) & 3748.6 ($\pm$ 15.62) & 0.2 ($\pm$ 0.40) & 2876.6 ($\pm$ 54.19) \\
        Reas., In-cont. (cat.), Dir. stim. & 0.2 ($\pm$ 0.40) & 3685.6 ($\pm$ 14.89) & 0.0 ($\pm$ 0.00) & 3388.0 ($\pm$ 61.31) & 22.6 ($\pm$ 2.58) & 2957.2 ($\pm$ 64.31) \\
        Reas., In-cont. (cat.), Dir. stim., Pers. & 0.6 ($\pm$ 1.20) & 3666.2 ($\pm$ 13.89) & 0.0 ($\pm$ 0.00) & 3666.4 ($\pm$ 38.76) & 0.6 ($\pm$ 0.80) & 2718.6 ($\pm$ 49.47) \\
        Reas., In-cont. (cat.), Pers. & 0.2 ($\pm$ 0.40) & 3674.2 ($\pm$ 22.96) & 0.0 ($\pm$ 0.00) & 3673.8 ($\pm$ 41.86) & 0.0 ($\pm$ 0.00) & 2929.8 ($\pm$ 59.57) \\
        Reas., In-cont. (rand.) & 0.6 ($\pm$ 0.80) & 3670.4 ($\pm$ 5.12) & 0.2 ($\pm$ 0.40) & 3633.8 ($\pm$ 36.64) & 0.0 ($\pm$ 0.00) & 3206.8 ($\pm$ 93.68) \\
        Reas., In-cont. (rand.), Pers. & 0.2 ($\pm$ 0.40) & 3659.0 ($\pm$ 9.88) & 0.0 ($\pm$ 0.00) & 3622.6 ($\pm$ 20.58) & 41.6 ($\pm$ 24.20) & 2728.2 ($\pm$ 175.78) \\
        Reas., In-cont. (sim.) & 0.2 ($\pm$ 0.40) & 3708.6 ($\pm$ 14.61) & 11.0 ($\pm$ 14.68) & 3655.2 ($\pm$ 25.59) & 0.2 ($\pm$ 0.40) & 3189.2 ($\pm$ 172.89) \\
        Reas., In-cont. (sim.), Pers. & 0.0 ($\pm$ 0.00) & 3687.4 ($\pm$ 21.56) & 16.0 ($\pm$ 15.74) & 3645.2 ($\pm$ 21.33) & 0.0 ($\pm$ 0.00) & 2582.4 ($\pm$ 62.88) \\
        Reas., Pers. & 0.2 ($\pm$ 0.40) & 3281.0 ($\pm$ 0.00) & 13.6 ($\pm$ 7.89) & 2727.0 ($\pm$ 0.00) & 2.0 ($\pm$ 1.67) & 3115.0 ($\pm$ 0.00) \\

        \bottomrule
    \end{tabular}

    \caption{Frequencies of how often each composition was chosen as optimal composition by our adaptive prompting approach per LLM on SBIC. Frequencies are averaged over five random seeds. Possible techniques for a composition  are a defintion (\textit{Def.}), a directional stimulus (\textit{Dir. stim.}), In-context examples chosen randomly (\textit{In-cont. (rand.)}), based on similarity (\textit{In-cont. (sim.)}) or based on their category (\textit{In-cont. (cat.)}), a persona (\textit{Pers.}), and reasoning steps (\textit{Reas.}). The \textit{Base Composition} consists of a task description and text input.}
    \label{tab:composition-frequencies-sbic}
\end{table*}

\begin{table*}
    \scriptsize
    \centering
    \setlength{\tabcolsep}{1.8pt}

    \begin{tabular}{lrrrrrr}
        \toprule
        & \multicolumn{2}{c}{\textbf{Mistral}} & \multicolumn{2}{c}{\textbf{Command-R}} & \multicolumn{2}{c}{\textbf{Llama 3}} \\
        \cmidrule(l){2-3} \cmidrule(l){4-5} \cmidrule(l){6-7}
        \textbf{Composition} & \multicolumn{1}{c}{\textbf{Frequency}} & \multicolumn{1}{c}{\textbf{Correct on train}} & \multicolumn{1}{c}{\textbf{Frequency}} & \multicolumn{1}{c}{\textbf{Correct on train}} & \multicolumn{1}{c}{\textbf{Frequency}} & \multicolumn{1}{c}{\textbf{Correct on train}} \\
        \midrule

        Base composition & 0.0 ($\pm$ 0.00) & 1529.0 ($\pm$ 0.00) & 0.0 ($\pm$ 0.00) & 1700.0 ($\pm$ 0.00) & 0.0 ($\pm$ 0.00) & 1224.0 ($\pm$ 0.00) \\
        Def. & 0.0 ($\pm$ 0.00) & 1540.0 ($\pm$ 0.00) & 0.0 ($\pm$ 0.00) & 1741.0 ($\pm$ 0.00) & 0.0 ($\pm$ 0.00) & 1326.0 ($\pm$ 0.00) \\
        Def., Dir. stim. & 0.0 ($\pm$ 0.00) & 1489.0 ($\pm$ 0.00) & 0.0 ($\pm$ 0.00) & 1663.0 ($\pm$ 0.00) & 0.0 ($\pm$ 0.00) & 1549.0 ($\pm$ 0.00) \\
        Def., Dir. stim., In-cont. (rand.) & 5.2 ($\pm$ 10.40) & 1447.2 ($\pm$ 22.27) & 5.2 ($\pm$ 10.40) & 1616.2 ($\pm$ 12.50) & 5.2 ($\pm$ 10.40) & 1494.4 ($\pm$ 70.63) \\
        Def., Dir. stim., In-cont. (rand.), Pers. & 0.0 ($\pm$ 0.00) & 1440.4 ($\pm$ 18.19) & 0.0 ($\pm$ 0.00) & 1638.4 ($\pm$ 12.11) & 0.0 ($\pm$ 0.00) & 1421.8 ($\pm$ 78.79) \\
        Def., Dir. stim., In-cont. (sim.) & 0.0 ($\pm$ 0.00) & 1508.4 ($\pm$ 22.17) & 0.0 ($\pm$ 0.00) & 1680.0 ($\pm$ 19.71) & 0.0 ($\pm$ 0.00) & 1513.0 ($\pm$ 71.25) \\
        Def., Dir. stim., In-cont. (sim.), Pers. & 0.0 ($\pm$ 0.00) & 1496.0 ($\pm$ 24.55) & 0.0 ($\pm$ 0.00) & 1692.8 ($\pm$ 13.66) & 0.0 ($\pm$ 0.00) & 1447.8 ($\pm$ 76.72) \\
        Def., Dir. stim., Pers. & 0.0 ($\pm$ 0.00) & 1511.0 ($\pm$ 0.00) & 0.0 ($\pm$ 0.00) & 1620.0 ($\pm$ 0.00) & 0.0 ($\pm$ 0.00) & 1660.0 ($\pm$ 0.00) \\
        Def., In-cont. (rand.) & 0.0 ($\pm$ 0.00) & 1529.0 ($\pm$ 29.82) & 0.0 ($\pm$ 0.00) & 1563.2 ($\pm$ 35.64) & 0.0 ($\pm$ 0.00) & 1601.2 ($\pm$ 42.71) \\
        Def., In-cont. (rand.), Pers. & 0.0 ($\pm$ 0.00) & 1521.8 ($\pm$ 29.74) & 0.0 ($\pm$ 0.00) & 1593.6 ($\pm$ 28.30) & 0.0 ($\pm$ 0.00) & 1542.6 ($\pm$ 37.98) \\
        Def., In-cont. (sim.) & 323.2 ($\pm$ 310.68) & 1597.4 ($\pm$ 23.65) & 323.2 ($\pm$ 310.68) & 1644.8 ($\pm$ 27.93) & 323.2 ($\pm$ 310.68) & 1617.0 ($\pm$ 16.30) \\
        Def., In-cont. (sim.), Pers. & 177.8 ($\pm$ 125.53) & 1592.2 ($\pm$ 20.23) & 177.8 ($\pm$ 125.53) & 1665.0 ($\pm$ 22.18) & 177.8 ($\pm$ 125.53) & 1538.4 ($\pm$ 11.41) \\
        Def., Pers. & 0.0 ($\pm$ 0.00) & 1558.0 ($\pm$ 0.00) & 0.0 ($\pm$ 0.00) & 1680.0 ($\pm$ 0.00) & 0.0 ($\pm$ 0.00) & 1205.0 ($\pm$ 0.00) \\
        Dir. stim. & 0.0 ($\pm$ 0.00) & 1463.0 ($\pm$ 0.00) & 0.0 ($\pm$ 0.00) & 1487.0 ($\pm$ 0.00) & 0.0 ($\pm$ 0.00) & 1402.0 ($\pm$ 0.00) \\
        Dir. stim., In-cont. (rand.) & 0.0 ($\pm$ 0.00) & 1424.4 ($\pm$ 23.10) & 0.0 ($\pm$ 0.00) & 1606.2 ($\pm$ 19.33) & 0.0 ($\pm$ 0.00) & 1471.0 ($\pm$ 67.20) \\
        Dir. stim., In-cont. (rand.), Pers. & 0.0 ($\pm$ 0.00) & 1419.2 ($\pm$ 20.54) & 0.0 ($\pm$ 0.00) & 1615.8 ($\pm$ 13.04) & 0.0 ($\pm$ 0.00) & 1431.0 ($\pm$ 70.65) \\
        Dir. stim., In-cont. (sim.) & 6.0 ($\pm$ 12.00) & 1494.6 ($\pm$ 18.75) & 6.0 ($\pm$ 12.00) & 1675.0 ($\pm$ 16.02) & 6.0 ($\pm$ 12.00) & 1508.4 ($\pm$ 63.90) \\
        Dir. stim., In-cont. (sim.), Pers. & 2.2 ($\pm$ 4.40) & 1473.8 ($\pm$ 24.51) & 2.2 ($\pm$ 4.40) & 1683.2 ($\pm$ 12.81) & 2.2 ($\pm$ 4.40) & 1464.2 ($\pm$ 69.65) \\
        Dir. stim., Pers. & 0.0 ($\pm$ 0.00) & 1531.0 ($\pm$ 0.00) & 0.0 ($\pm$ 0.00) & 1479.0 ($\pm$ 0.00) & 0.0 ($\pm$ 0.00) & 1278.0 ($\pm$ 0.00) \\
        In-cont. (cat.) & 508.6 ($\pm$ 98.07) & 1568.2 ($\pm$ 13.26) & 508.6 ($\pm$ 98.07) & 1518.4 ($\pm$ 41.74) & 508.6 ($\pm$ 98.07) & 1644.6 ($\pm$ 16.26) \\
        In-cont. (cat.), Def. & 155.6 ($\pm$ 160.16) & 1546.4 ($\pm$ 8.80) & 155.6 ($\pm$ 160.16) & 1515.2 ($\pm$ 33.78) & 155.6 ($\pm$ 160.16) & 1651.6 ($\pm$ 17.35) \\
        In-cont. (cat.), Def., Dir. stim. & 0.0 ($\pm$ 0.00) & 1437.8 ($\pm$ 8.01) & 0.0 ($\pm$ 0.00) & 1601.2 ($\pm$ 15.47) & 0.0 ($\pm$ 0.00) & 1663.6 ($\pm$ 13.37) \\
        In-cont. (cat.), Def., Dir. stim., Pers. & 0.0 ($\pm$ 0.00) & 1430.0 ($\pm$ 12.13) & 0.0 ($\pm$ 0.00) & 1609.6 ($\pm$ 11.48) & 0.0 ($\pm$ 0.00) & 1603.0 ($\pm$ 4.15) \\
        In-cont. (cat.), Def., Pers. & 75.4 ($\pm$ 149.80) & 1522.4 ($\pm$ 11.81) & 75.4 ($\pm$ 149.80) & 1537.4 ($\pm$ 28.88) & 75.4 ($\pm$ 149.80) & 1576.0 ($\pm$ 25.10) \\
        In-cont. (cat.), Dir. stim. & 0.2 ($\pm$ 0.40) & 1428.2 ($\pm$ 12.19) & 0.2 ($\pm$ 0.40) & 1604.6 ($\pm$ 21.40) & 0.2 ($\pm$ 0.40) & 1652.6 ($\pm$ 12.64) \\
        In-cont. (cat.), Dir. stim., Pers. & 0.0 ($\pm$ 0.00) & 1410.8 ($\pm$ 13.73) & 0.0 ($\pm$ 0.00) & 1602.2 ($\pm$ 15.71) & 0.0 ($\pm$ 0.00) & 1608.6 ($\pm$ 7.42) \\
        In-cont. (cat.), Pers. & 87.0 ($\pm$ 124.03) & 1527.0 ($\pm$ 15.43) & 87.0 ($\pm$ 124.03) & 1529.4 ($\pm$ 27.17) & 87.0 ($\pm$ 124.03) & 1575.4 ($\pm$ 23.64) \\
        In-cont. (rand.) & 0.0 ($\pm$ 0.00) & 1532.4 ($\pm$ 29.51) & 0.0 ($\pm$ 0.00) & 1535.4 ($\pm$ 46.43) & 0.0 ($\pm$ 0.00) & 1570.0 ($\pm$ 46.12) \\
        In-cont. (rand.), Pers. & 0.0 ($\pm$ 0.00) & 1504.2 ($\pm$ 28.24) & 0.0 ($\pm$ 0.00) & 1560.6 ($\pm$ 33.09) & 0.0 ($\pm$ 0.00) & 1524.6 ($\pm$ 37.03) \\
        In-cont. (sim.) & 384.8 ($\pm$ 265.66) & 1596.4 ($\pm$ 25.35) & 384.8 ($\pm$ 265.66) & 1618.4 ($\pm$ 41.35) & 384.8 ($\pm$ 265.66) & 1601.4 ($\pm$ 21.85) \\
        In-cont. (sim.), Pers. & 2.8 ($\pm$ 2.64) & 1576.0 ($\pm$ 23.48) & 2.8 ($\pm$ 2.64) & 1640.2 ($\pm$ 25.36) & 2.8 ($\pm$ 2.64) & 1544.0 ($\pm$ 20.32) \\
        Pers. & 0.0 ($\pm$ 0.00) & 1547.0 ($\pm$ 0.00) & 0.0 ($\pm$ 0.00) & 1666.0 ($\pm$ 0.00) & 0.0 ($\pm$ 0.00) & 909.0 ($\pm$ 0.00) \\
        Reas., Base composition & 7.8 ($\pm$ 14.15) & 1617.0 ($\pm$ 0.00) & 7.8 ($\pm$ 14.15) & 1756.0 ($\pm$ 0.00) & 7.8 ($\pm$ 14.15) & 1313.0 ($\pm$ 0.00) \\
        Reas., Def. & 9.2 ($\pm$ 10.98) & 1614.0 ($\pm$ 0.00) & 9.2 ($\pm$ 10.98) & 1718.0 ($\pm$ 0.00) & 9.2 ($\pm$ 10.98) & 1386.0 ($\pm$ 0.00) \\
        Reas., Def., Dir. stim. & 0.0 ($\pm$ 0.00) & 1437.0 ($\pm$ 0.00) & 0.0 ($\pm$ 0.00) & 1629.0 ($\pm$ 0.00) & 0.0 ($\pm$ 0.00) & 1353.0 ($\pm$ 0.00) \\
        Reas., Def., Dir. stim., In-cont. (rand.) & 0.0 ($\pm$ 0.00) & 1506.2 ($\pm$ 8.86) & 0.0 ($\pm$ 0.00) & 1758.8 ($\pm$ 10.42) & 0.0 ($\pm$ 0.00) & 1168.8 ($\pm$ 56.00) \\
        Reas., Def., Dir. stim., In-cont. (rand.), Pers. & 0.4 ($\pm$ 0.80) & 1491.4 ($\pm$ 7.06) & 0.4 ($\pm$ 0.80) & 1769.2 ($\pm$ 4.49) & 0.4 ($\pm$ 0.80) & 794.2 ($\pm$ 154.33) \\
        Reas., Def., Dir. stim., In-cont. (sim.) & 0.0 ($\pm$ 0.00) & 1457.2 ($\pm$ 11.16) & 0.0 ($\pm$ 0.00) & 1750.6 ($\pm$ 5.68) & 0.0 ($\pm$ 0.00) & 1233.8 ($\pm$ 76.43) \\
        Reas., Def., Dir. stim., In-cont. (sim.), Pers. & 0.0 ($\pm$ 0.00) & 1446.8 ($\pm$ 19.23) & 0.0 ($\pm$ 0.00) & 1770.8 ($\pm$ 11.36) & 0.0 ($\pm$ 0.00) & 853.6 ($\pm$ 160.12) \\
        Reas., Def., Dir. stim., Pers. & 0.0 ($\pm$ 0.00) & 1428.0 ($\pm$ 0.00) & 0.0 ($\pm$ 0.00) & 1679.0 ($\pm$ 0.00) & 0.0 ($\pm$ 0.00) & 463.0 ($\pm$ 0.00) \\
        Reas., Def., In-cont. (rand.) & 0.2 ($\pm$ 0.40) & 1539.0 ($\pm$ 12.98) & 0.2 ($\pm$ 0.40) & 1761.4 ($\pm$ 13.00) & 0.2 ($\pm$ 0.40) & 1465.0 ($\pm$ 47.34) \\
        Reas., Def., In-cont. (rand.), Pers. & 0.0 ($\pm$ 0.00) & 1527.0 ($\pm$ 14.44) & 0.0 ($\pm$ 0.00) & 1770.2 ($\pm$ 10.24) & 0.0 ($\pm$ 0.00) & 1193.0 ($\pm$ 187.78) \\
        Reas., Def., In-cont. (sim.) & 0.0 ($\pm$ 0.00) & 1489.6 ($\pm$ 9.73) & 0.0 ($\pm$ 0.00) & 1745.6 ($\pm$ 7.47) & 0.0 ($\pm$ 0.00) & 1446.6 ($\pm$ 36.42) \\
        Reas., Def., In-cont. (sim.), Pers. & 0.0 ($\pm$ 0.00) & 1468.6 ($\pm$ 7.84) & 0.0 ($\pm$ 0.00) & 1767.8 ($\pm$ 8.01) & 0.0 ($\pm$ 0.00) & 1058.0 ($\pm$ 269.41) \\
        Reas., Def., Pers. & 0.0 ($\pm$ 0.00) & 1612.0 ($\pm$ 0.00) & 0.0 ($\pm$ 0.00) & 1721.0 ($\pm$ 0.00) & 0.0 ($\pm$ 0.00) & 952.0 ($\pm$ 0.00) \\
        Reas., Dir. stim. & 0.0 ($\pm$ 0.00) & 1446.0 ($\pm$ 0.00) & 0.0 ($\pm$ 0.00) & 1642.0 ($\pm$ 0.00) & 0.0 ($\pm$ 0.00) & 1098.0 ($\pm$ 0.00) \\
        Reas., Dir. stim., In-cont. (rand.) & 0.0 ($\pm$ 0.00) & 1481.4 ($\pm$ 12.34) & 0.0 ($\pm$ 0.00) & 1759.0 ($\pm$ 5.83) & 0.0 ($\pm$ 0.00) & 1178.2 ($\pm$ 55.03) \\
        Reas., Dir. stim., In-cont. (rand.), Pers. & 1.8 ($\pm$ 3.60) & 1472.4 ($\pm$ 13.17) & 1.8 ($\pm$ 3.60) & 1763.0 ($\pm$ 12.13) & 1.8 ($\pm$ 3.60) & 861.2 ($\pm$ 533.99) \\
        Reas., Dir. stim., In-cont. (sim.) & 0.0 ($\pm$ 0.00) & 1452.8 ($\pm$ 8.63) & 0.0 ($\pm$ 0.00) & 1743.4 ($\pm$ 11.57) & 0.0 ($\pm$ 0.00) & 1261.6 ($\pm$ 62.89) \\
        Reas., Dir. stim., In-cont. (sim.), Pers. & 0.0 ($\pm$ 0.00) & 1431.4 ($\pm$ 10.25) & 0.0 ($\pm$ 0.00) & 1762.4 ($\pm$ 8.52) & 0.0 ($\pm$ 0.00) & 841.6 ($\pm$ 498.75) \\
        Reas., Dir. stim., Pers. & 0.0 ($\pm$ 0.00) & 1410.0 ($\pm$ 0.00) & 0.0 ($\pm$ 0.00) & 1696.0 ($\pm$ 0.00) & 0.0 ($\pm$ 0.00) & 881.0 ($\pm$ 0.00) \\
        Reas., In-cont. (cat.) & 1.4 ($\pm$ 2.80) & 1484.6 ($\pm$ 13.95) & 1.4 ($\pm$ 2.80) & 1769.2 ($\pm$ 10.89) & 1.4 ($\pm$ 2.80) & 1693.2 ($\pm$ 14.82) \\
        Reas., In-cont. (cat.), Def. & 0.0 ($\pm$ 0.00) & 1481.6 ($\pm$ 14.24) & 0.0 ($\pm$ 0.00) & 1756.8 ($\pm$ 13.63) & 0.0 ($\pm$ 0.00) & 1700.2 ($\pm$ 23.14) \\
        Reas., In-cont. (cat.), Def., Dir. stim. & 0.0 ($\pm$ 0.00) & 1448.2 ($\pm$ 8.45) & 0.0 ($\pm$ 0.00) & 1750.6 ($\pm$ 9.89) & 0.0 ($\pm$ 0.00) & 1723.6 ($\pm$ 10.37) \\
        Reas., In-cont. (cat.), Def., Dir. stim., Pers. & 0.0 ($\pm$ 0.00) & 1407.0 ($\pm$ 7.97) & 0.0 ($\pm$ 0.00) & 1767.4 ($\pm$ 6.09) & 0.0 ($\pm$ 0.00) & 1636.2 ($\pm$ 5.04) \\
        Reas., In-cont. (cat.), Def., Pers. & 0.0 ($\pm$ 0.00) & 1447.6 ($\pm$ 8.87) & 0.0 ($\pm$ 0.00) & 1770.0 ($\pm$ 13.68) & 0.0 ($\pm$ 0.00) & 1617.0 ($\pm$ 48.92) \\
        Reas., In-cont. (cat.), Dir. stim. & 0.0 ($\pm$ 0.00) & 1458.6 ($\pm$ 14.64) & 0.0 ($\pm$ 0.00) & 1753.6 ($\pm$ 10.40) & 0.0 ($\pm$ 0.00) & 1737.0 ($\pm$ 10.49) \\
        Reas., In-cont. (cat.), Dir. stim., Pers. & 0.0 ($\pm$ 0.00) & 1418.2 ($\pm$ 7.19) & 0.0 ($\pm$ 0.00) & 1776.0 ($\pm$ 4.24) & 0.0 ($\pm$ 0.00) & 1720.2 ($\pm$ 5.64) \\
        Reas., In-cont. (cat.), Pers. & 0.0 ($\pm$ 0.00) & 1455.4 ($\pm$ 14.69) & 0.0 ($\pm$ 0.00) & 1785.8 ($\pm$ 9.28) & 0.0 ($\pm$ 0.00) & 1548.2 ($\pm$ 165.73) \\
        Reas., In-cont. (rand.) & 18.4 ($\pm$ 36.80) & 1511.6 ($\pm$ 11.36) & 18.4 ($\pm$ 36.80) & 1754.4 ($\pm$ 3.61) & 18.4 ($\pm$ 36.80) & 1371.8 ($\pm$ 21.16) \\
        Reas., In-cont. (rand.), Pers. & 0.0 ($\pm$ 0.00) & 1501.4 ($\pm$ 9.65) & 0.0 ($\pm$ 0.00) & 1778.4 ($\pm$ 11.00) & 0.0 ($\pm$ 0.00) & 1002.4 ($\pm$ 204.50) \\
        Reas., In-cont. (sim.) & 0.0 ($\pm$ 0.00) & 1476.2 ($\pm$ 10.46) & 0.0 ($\pm$ 0.00) & 1744.6 ($\pm$ 10.44) & 0.0 ($\pm$ 0.00) & 1362.2 ($\pm$ 44.01) \\
        Reas., In-cont. (sim.), Pers. & 0.0 ($\pm$ 0.00) & 1452.6 ($\pm$ 7.39) & 0.0 ($\pm$ 0.00) & 1759.6 ($\pm$ 12.99) & 0.0 ($\pm$ 0.00) & 1073.2 ($\pm$ 192.26) \\
        Reas., Pers. & 171.0 ($\pm$ 189.08) & 1645.0 ($\pm$ 0.00) & 171.0 ($\pm$ 189.08) & 1740.0 ($\pm$ 0.00) & 171.0 ($\pm$ 189.08) & 968.0 ($\pm$ 0.00) \\

        \bottomrule
    \end{tabular}

    \caption{Frequencies of how often each composition was chosen as optimal composition by our adaptive prompting approach per LLM on CobraFrames. Frequencies are averaged over five random seeds. Possible techniques for a composition are a defintion (\textit{Def.}), a directional stimulus (\textit{Dir. stim.}), In-context examples chosen randomly (\textit{In-cont. (rand.)}), based on similarity (\textit{In-cont. (sim.)}) or based on their category (\textit{In-cont. (cat.)}), a persona (\textit{Pers.}), and reasoning steps (\textit{Reas.}). The \textit{Base Composition} consists of a task description and text input.}
    \label{tab:composition-frequencies-cobra}
\end{table*}

\begin{table*}
    \small
    \centering
    \begin{tabular}{lrrr}
        \toprule
        \textbf{Composition} & \textbf{Mistral} & \textbf{Command-R} & \textbf{Llama 3} \\
        \midrule
        Base composition & 0.711 & 0.462 & 0.575 \\
        Definition & 0.716 & 0.527 & 0.637 \\
        Definition, Dir. stimulus & 0.672 & 0.498 & 0.544 \\
        Definition, Dir. stimulus, In-context (random) & 0.636 & 0.592 & 0.734 \\
        Definition, Dir. stimulus, In-context (random), Persona & 0.629 & 0.588 & 0.735 \\
        Definition, Dir. stimulus, In-context (similar) & 0.766 & 0.610 & 0.798 \\
        Definition, Dir. stimulus, In-context (similar), Persona & 0.767 & 0.582 & 0.802 \\
        Definition, Dir. stimulus, Persona & 0.676 & 0.542 & 0.483 \\
        Definition, In-context (random) & 0.671 & 0.667 & 0.736 \\
        Definition, In-context (random), Persona & 0.660 & 0.667 & 0.748 \\
        Definition, In-context (similar) & 0.775 & 0.683 & 0.800 \\
        Definition, In-context (similar), Persona & 0.776 & 0.671 & 0.817 \\
        Definition, Persona & 0.710 & 0.591 & 0.502 \\
        Dir. stimulus & 0.662 & 0.584 & 0.566 \\
        Dir. stimulus, In-context (random) & 0.634 & 0.590 & 0.726 \\
        Dir. stimulus, In-context (random), Persona & 0.632 & 0.598 & 0.733 \\
        Dir. stimulus, In-context (similar) & 0.759 & 0.600 & 0.795 \\
        Dir. stimulus, In-context (similar), Persona & 0.764 & 0.579 & 0.799 \\
        Dir. stimulus, Persona & 0.654 & 0.602 & 0.508 \\
        In-context (category) & 0.681 & 0.675 & 0.739 \\
        In-context (category), Definition & 0.681 & 0.663 & 0.760 \\
        In-context (category), Definition, Dir. stimulus & 0.652 & 0.571 & 0.715 \\
        In-context (category), Definition, Dir. stimulus, Persona & 0.645 & 0.579 & 0.728 \\
        In-context (category), Definition, Persona & 0.687 & 0.669 & 0.759 \\
        In-context (category), Dir. stimulus & 0.650 & 0.580 & 0.705 \\
        In-context (category), Dir. stimulus, Persona & 0.637 & 0.594 & 0.725 \\
        In-context (category), Persona & 0.665 & 0.685 & 0.738 \\
        In-context (random) & 0.665 & 0.674 & 0.725 \\
        In-context (random), Persona & 0.652 & 0.677 & 0.736 \\
        In-context (similar) & 0.761 & 0.701 & 0.798 \\
        In-context (similar), Persona & 0.763 & 0.706 & 0.814 \\
        Persona & 0.698 & 0.546 & 0.539 \\
        Reasoning steps & 0.697 & 0.509 & 0.610 \\
        Reasoning steps, Definition & 0.722 & 0.491 & 0.693 \\
        Reasoning steps, Definition, Dir. stimulus & 0.688 & 0.520 & 0.584 \\
        Reasoning steps, Definition, Dir. stimulus, In-context (random) & 0.705 & 0.609 & 0.661 \\
        Reasoning steps, Definition, Dir. stimulus, In-context (random), Persona & 0.705 & 0.652 & 0.495 \\
        Reasoning steps, Definition, Dir. stimulus, In-context (similar) & 0.797 & 0.596 & 0.776 \\
        Reasoning steps, Definition, Dir. stimulus, In-context (similar), Persona & 0.795 & 0.650 & 0.608 \\
        Reasoning steps, Definition, Dir. stimulus, Persona & 0.670 & 0.535 & 0.590 \\
        Reasoning steps, Definition, In-context (random) & 0.719 & 0.580 & 0.755 \\
        Reasoning steps, Definition, In-context (random), Persona & 0.716 & 0.642 & 0.561 \\
        Reasoning steps, Definition, In-context (similar) & 0.800 & 0.570 & 0.806 \\
        Reasoning steps, Definition, In-context (similar), Persona & 0.798 & 0.629 & 0.501 \\
        Reasoning steps, Definition, Persona & 0.692 & 0.521 & 0.635 \\
        Reasoning steps, Dir. stimulus & 0.646 & 0.442 & 0.602 \\
        Reasoning steps, Dir. stimulus, In-context (random) & 0.701 & 0.630 & 0.738 \\
        Reasoning steps, Dir. stimulus, In-context (random), Persona & 0.703 & 0.640 & 0.540 \\
        Reasoning steps, Dir. stimulus, In-context (similar) & 0.791 & 0.603 & 0.776 \\
        Reasoning steps, Dir. stimulus, In-context (similar), Persona & 0.791 & 0.677 & 0.596 \\
        Reasoning steps, Dir. stimulus, Persona & 0.658 & 0.465 & 0.606 \\
        Reasoning steps, In-context (category) & 0.705 & 0.440 & 0.728 \\
        Reasoning steps, In-context (category), Definition & 0.703 & 0.414 & 0.742 \\
        Reasoning steps, In-context (category), Definition, Dir. stimulus & 0.687 & 0.407 & 0.702 \\
        Reasoning steps, In-context (category), Definition, Dir. stimulus, Persona & 0.682 & 0.582 & 0.583 \\
        Reasoning steps, In-context (category), Definition, Persona & 0.697 & 0.526 & 0.641 \\
        Reasoning steps, In-context (category), Dir. stimulus & 0.696 & 0.419 & 0.701 \\
        Reasoning steps, In-context (category), Dir. stimulus, Persona & 0.691 & 0.578 & 0.582 \\
        Reasoning steps, In-context (category), Persona & 0.698 & 0.538 & 0.597 \\
        Reasoning steps, In-context (random) & 0.705 & 0.610 & 0.768 \\
        Reasoning steps, In-context (random), Persona & 0.709 & 0.665 & 0.543 \\
        Reasoning steps, In-context (similar) & 0.791 & 0.538 & 0.806 \\
        Reasoning steps, In-context (similar), Persona & 0.790 & 0.664 & 0.518 \\
        Reasoning steps, Persona & 0.693 & 0.506 & 0.659 \\
        \bottomrule
    \end{tabular}

    \caption{Macro F$_1$-score of all compositions across models on Stereoset.}
    \label{tab:compositions-performance-stereoset}
\end{table*}

\begin{table*}
    \small
    \centering
    \begin{tabular}{lrrr}
        \toprule
        \textbf{Composition} & \textbf{Mistral} & \textbf{Command-R} & \textbf{Llama 3} \\
        \midrule
        Base composition & 0.702 & 0.470 & 0.651 \\
        Definition & 0.740 & 0.554 & 0.788 \\
        Definition, Dir. stimulus & 0.742 & 0.425 & 0.741 \\
        Definition, Dir. stimulus, In-context (random) & 0.792 & 0.773 & 0.817 \\
        Definition, Dir. stimulus, In-context (random), Persona & 0.791 & 0.772 & 0.821 \\
        Definition, Dir. stimulus, In-context (similar) & 0.758 & 0.770 & 0.822 \\
        Definition, Dir. stimulus, In-context (similar), Persona & 0.756 & 0.764 & 0.820 \\
        Definition, Dir. stimulus, Persona & 0.732 & 0.447 & 0.683 \\
        Definition, In-context (random) & 0.769 & 0.787 & 0.826 \\
        Definition, In-context (random), Persona & 0.765 & 0.788 & 0.831 \\
        Definition, In-context (similar) & 0.740 & 0.770 & 0.821 \\
        Definition, In-context (similar), Persona & 0.734 & 0.766 & 0.825 \\
        Definition, Persona & 0.727 & 0.596 & 0.771 \\
        Dir. stimulus & 0.725 & 0.410 & 0.542 \\
        Dir. stimulus, In-context (random) & 0.780 & 0.768 & 0.820 \\
        Dir. stimulus, In-context (random), Persona & 0.777 & 0.760 & 0.824 \\
        Dir. stimulus, In-context (similar) & 0.749 & 0.760 & 0.822 \\
        Dir. stimulus, In-context (similar), Persona & 0.746 & 0.749 & 0.818 \\
        Dir. stimulus, Persona & 0.720 & 0.459 & 0.646 \\
        In-context (category) & 0.737 & 0.733 & 0.806 \\
        In-context (category), Definition & 0.762 & 0.772 & 0.810 \\
        In-context (category), Definition, Dir. stimulus & 0.781 & 0.765 & 0.783 \\
        In-context (category), Definition, Dir. stimulus, Persona & 0.783 & 0.760 & 0.794 \\
        In-context (category), Definition, Persona & 0.760 & 0.767 & 0.821 \\
        In-context (category), Dir. stimulus & 0.764 & 0.745 & 0.778 \\
        In-context (category), Dir. stimulus, Persona & 0.762 & 0.737 & 0.789 \\
        In-context (category), Persona & 0.737 & 0.728 & 0.817 \\
        In-context (random) & 0.747 & 0.763 & 0.825 \\
        In-context (random), Persona & 0.744 & 0.762 & 0.829 \\
        In-context (similar) & 0.716 & 0.740 & 0.816 \\
        In-context (similar), Persona & 0.712 & 0.729 & 0.822 \\
        Persona & 0.703 & 0.512 & 0.710 \\
        Reasoning steps & 0.656 & 0.436 & 0.621 \\
        Reasoning steps, Definition & 0.697 & 0.494 & 0.647 \\
        Reasoning steps, Definition, Dir. stimulus & 0.704 & 0.473 & 0.684 \\
        Reasoning steps, Definition, Dir. stimulus, In-context (random) & 0.776 & 0.743 & 0.663 \\
        Reasoning steps, Definition, Dir. stimulus, In-context (random), Persona & 0.772 & 0.724 & 0.502 \\
        Reasoning steps, Definition, Dir. stimulus, In-context (similar) & 0.771 & 0.730 & 0.604 \\
        Reasoning steps, Definition, Dir. stimulus, In-context (similar), Persona & 0.770 & 0.712 & 0.469 \\
        Reasoning steps, Definition, Dir. stimulus, Persona & 0.711 & 0.429 & 0.723 \\
        Reasoning steps, Definition, In-context (random) & 0.778 & 0.754 & 0.677 \\
        Reasoning steps, Definition, In-context (random), Persona & 0.775 & 0.747 & 0.505 \\
        Reasoning steps, Definition, In-context (similar) & 0.772 & 0.735 & 0.618 \\
        Reasoning steps, Definition, In-context (similar), Persona & 0.769 & 0.729 & 0.430 \\
        Reasoning steps, Definition, Persona & 0.708 & 0.453 & 0.699 \\
        Reasoning steps, Dir. stimulus & 0.687 & 0.484 & 0.630 \\
        Reasoning steps, Dir. stimulus, In-context (random) & 0.768 & 0.719 & 0.611 \\
        Reasoning steps, Dir. stimulus, In-context (random), Persona & 0.767 & 0.702 & 0.435 \\
        Reasoning steps, Dir. stimulus, In-context (similar) & 0.768 & 0.705 & 0.572 \\
        Reasoning steps, Dir. stimulus, In-context (similar), Persona & 0.765 & 0.688 & 0.412 \\
        Reasoning steps, Dir. stimulus, Persona & 0.712 & 0.420 & 0.699 \\
        Reasoning steps, In-context (category) & 0.761 & 0.581 & 0.647 \\
        Reasoning steps, In-context (category), Definition & 0.773 & 0.683 & 0.679 \\
        Reasoning steps, In-context (category), Definition, Dir. stimulus & 0.772 & 0.672 & 0.668 \\
        Reasoning steps, In-context (category), Definition, Dir. stimulus, Persona & 0.771 & 0.751 & 0.556 \\
        Reasoning steps, In-context (category), Definition, Persona & 0.773 & 0.765 & 0.548 \\
        Reasoning steps, In-context (category), Dir. stimulus & 0.765 & 0.647 & 0.584 \\
        Reasoning steps, In-context (category), Dir. stimulus, Persona & 0.764 & 0.731 & 0.534 \\
        Reasoning steps, In-context (category), Persona & 0.760 & 0.739 & 0.570 \\
        Reasoning steps, In-context (random) & 0.769 & 0.729 & 0.694 \\
        Reasoning steps, In-context (random), Persona & 0.769 & 0.725 & 0.488 \\
        Reasoning steps, In-context (similar) & 0.767 & 0.720 & 0.669 \\
        Reasoning steps, In-context (similar), Persona & 0.765 & 0.713 & 0.432 \\
        Reasoning steps, Persona & 0.678 & 0.408 & 0.668 \\
        \bottomrule
    \end{tabular}

    \caption{Macro F$_1$-score of all compositions across models on SBIC.}
    \label{tab:compositions-performance-sbic}
\end{table*}

\begin{table*}
    \small
    \centering
    \begin{tabular}{lrrr}
        \toprule
        \textbf{Composition} & \textbf{Mistral} & \textbf{Command-R} & \textbf{Llama 3} \\
        \midrule
        Base composition & 0.449 & 0.535 & 0.461 \\
        Definition & 0.485 & 0.575 & 0.497 \\
        Definition, Dir. stimulus & 0.466 & 0.544 & 0.431 \\
        Definition, Dir. stimulus, In-context (random) & 0.503 & 0.554 & 0.521 \\
        Definition, Dir. stimulus, In-context (random), Persona & 0.498 & 0.561 & 0.484 \\
        Definition, Dir. stimulus, In-context (similar) & 0.604 & 0.613 & 0.574 \\
        Definition, Dir. stimulus, In-context (similar), Persona & 0.602 & 0.611 & 0.554 \\
        Definition, Dir. stimulus, Persona & 0.462 & 0.518 & 0.498 \\
        Definition, In-context (random) & 0.533 & 0.534 & 0.576 \\
        Definition, In-context (random), Persona & 0.523 & 0.535 & 0.560 \\
        Definition, In-context (similar) & 0.604 & 0.580 & 0.594 \\
        Definition, In-context (similar), Persona & 0.598 & 0.587 & 0.582 \\
        Definition, Persona & 0.478 & 0.571 & 0.452 \\
        Dir. stimulus & 0.422 & 0.438 & 0.340 \\
        Dir. stimulus, In-context (random) & 0.497 & 0.546 & 0.519 \\
        Dir. stimulus, In-context (random), Persona & 0.489 & 0.557 & 0.493 \\
        Dir. stimulus, In-context (similar) & 0.602 & 0.615 & 0.591 \\
        Dir. stimulus, In-context (similar), Persona & 0.594 & 0.610 & 0.569 \\
        Dir. stimulus, Persona & 0.451 & 0.461 & 0.319 \\
        In-context (category) & 0.547 & 0.499 & 0.599 \\
        In-context (category), Definition & 0.537 & 0.489 & 0.598 \\
        In-context (category), Definition, Dir. stimulus & 0.493 & 0.540 & 0.599 \\
        In-context (category), Definition, Dir. stimulus, Persona & 0.487 & 0.542 & 0.573 \\
        In-context (category), Definition, Persona & 0.526 & 0.496 & 0.573 \\
        In-context (category), Dir. stimulus & 0.491 & 0.545 & 0.597 \\
        In-context (category), Dir. stimulus, Persona & 0.477 & 0.547 & 0.576 \\
        In-context (category), Persona & 0.527 & 0.505 & 0.578 \\
        In-context (random) & 0.537 & 0.530 & 0.566 \\
        In-context (random), Persona & 0.523 & 0.534 & 0.554 \\
        In-context (similar) & 0.604 & 0.588 & 0.605 \\
        In-context (similar), Persona & 0.597 & 0.590 & 0.593 \\
        Persona & 0.450 & 0.528 & 0.362 \\
        Reasoning steps & 0.535 & 0.589 & 0.417 \\
        Reasoning steps, Definition & 0.548 & 0.584 & 0.452 \\
        Reasoning steps, Definition, Dir. stimulus & 0.474 & 0.545 & 0.435 \\
        Reasoning steps, Definition, Dir. stimulus, In-context (random) & 0.523 & 0.646 & 0.415 \\
        Reasoning steps, Definition, Dir. stimulus, In-context (random), Persona & 0.515 & 0.645 & 0.318 \\
        Reasoning steps, Definition, Dir. stimulus, In-context (similar) & 0.540 & 0.646 & 0.498 \\
        Reasoning steps, Definition, Dir. stimulus, In-context (similar), Persona & 0.532 & 0.641 & 0.376 \\
        Reasoning steps, Definition, Dir. stimulus, Persona & 0.446 & 0.546 & 0.117 \\
        Reasoning steps, Definition, In-context (random) & 0.544 & 0.654 & 0.476 \\
        Reasoning steps, Definition, In-context (random), Persona & 0.536 & 0.650 & 0.386 \\
        Reasoning steps, Definition, In-context (similar) & 0.549 & 0.639 & 0.524 \\
        Reasoning steps, Definition, In-context (similar), Persona & 0.542 & 0.651 & 0.405 \\
        Reasoning steps, Definition, Persona & 0.533 & 0.560 & 0.284 \\
        Reasoning steps, Dir. stimulus & 0.452 & 0.559 & 0.350 \\
        Reasoning steps, Dir. stimulus, In-context (random) & 0.513 & 0.646 & 0.423 \\
        Reasoning steps, Dir. stimulus, In-context (random), Persona & 0.506 & 0.642 & 0.328 \\
        Reasoning steps, Dir. stimulus, In-context (similar) & 0.542 & 0.668 & 0.496 \\
        Reasoning steps, Dir. stimulus, In-context (similar), Persona & 0.532 & 0.660 & 0.369 \\
        Reasoning steps, Dir. stimulus, Persona & 0.432 & 0.571 & 0.251 \\
        Reasoning steps, In-context (category) & 0.521 & 0.643 & 0.533 \\
        Reasoning steps, In-context (category), Definition & 0.522 & 0.633 & 0.543 \\
        Reasoning steps, In-context (category), Definition, Dir. stimulus & 0.514 & 0.627 & 0.535 \\
        Reasoning steps, In-context (category), Definition, Dir. stimulus, Persona & 0.499 & 0.633 & 0.536 \\
        Reasoning steps, In-context (category), Definition, Persona & 0.501 & 0.631 & 0.524 \\
        Reasoning steps, In-context (category), Dir. stimulus & 0.512 & 0.639 & 0.535 \\
        Reasoning steps, In-context (category), Dir. stimulus, Persona & 0.497 & 0.651 & 0.548 \\
        Reasoning steps, In-context (category), Persona & 0.504 & 0.648 & 0.527 \\
        Reasoning steps, In-context (random) & 0.532 & 0.642 & 0.479 \\
        Reasoning steps, In-context (random), Persona & 0.530 & 0.644 & 0.369 \\
        Reasoning steps, In-context (similar) & 0.547 & 0.663 & 0.520 \\
        Reasoning steps, In-context (similar), Persona & 0.544 & 0.663 & 0.349 \\
        Reasoning steps, Persona & 0.531 & 0.610 & 0.302 \\
        \bottomrule
    \end{tabular}

    \caption{Macro F$_1$-score of all compositions across models on CobraFrames.}
    \label{tab:compositions-performance-cobra}
\end{table*}

\end{document}